\documentclass[preprint,12pt,3p,authoryear]{elsarticle}

\newcommand{\beginsupplement}{%
        \setcounter{table}{0}
        \renewcommand{\thetable}{S\arabic{table}}%
        \setcounter{figure}{0}
        \renewcommand{\thefigure}{S\arabic{figure}}%
     }

\usepackage{graphbox,graphicx}
\usepackage{amssymb}
\usepackage{gensymb}

\usepackage{lineno}

\usepackage{gensymb}
\usepackage[binary-units]{siunitx}
\usepackage{amsmath}
\usepackage{todonotes}
\usepackage{blkarray}
\usepackage{rotating}
\usepackage{fullpage}
\usepackage{times}
\usepackage{fancyhdr,graphicx,amsmath,amssymb}
\usepackage[ruled,vlined]{algorithm2e}
\usepackage{algorithmic}
\usepackage{url}

\journal{ISPRS J. Photogram. Remote Sens.}

\begin{document}

\begin{frontmatter}


\title{Classification of structural building damage grades from multi-temporal photogrammetric point clouds using a machine learning model trained on virtual laser scanning data}



\author[a]{Vivien Zahs}
\author[a]{Katharina Anders}
\author[b]{Julia Kohns}
\author[b]{Alexander Stark}
\author[a,c,d]{Bernhard Höfle}
\address[a]{3D Geospatial Data Processing (3DGeo) Research Group, Institute of Geography, Heidelberg University, Germany}
\address[b]{Institute of Concrete Structures and Building Materials, Karlsruhe Institute of Technology (KIT), Germany}
\address[c]{Interdisciplinary Center for Scientific Computing (IWR), Heidelberg University, Germany}
\address[d]{Heidelberg Center for the Environment (HCE), Heidelberg University, Germany}

\begin{abstract}
Timely and reliable information on earthquake-induced building damage plays a critical role for the effective planning of rescue and remediation actions. Automatic damage assessment based on the analysis of UAV-derived 3D point clouds, e.g., from photogrammetry or laser scanning, can provide fast and objective information on the damage situation within few hours. However, the assessment of different damage grades, beyond binary damage detection, is a challenging task due to the large variety in possible damage characteristics and limited transferability of existing methods to other geographic regions or data from different sources. 
We present a novel approach to automatically assess multi-class structural building damage
from real-world multi-temporal point clouds using a machine learning model
trained on virtual laser scanning (VLS) data. 
In our approach we (1)~identify object-specific change features, (2)~coarsely identify changed and unchanged building parts using a k-means clustering approach, (3)~train a random forest machine learning model with VLS data based on object-specific change features, and (4)~use the classifier to assess building damage grades in real-world point clouds from photogrammetry-based dense image matching (DIM). 
We evaluate classifiers trained on VLS point clouds and on real-world DIM point clouds with respect to their capacity to classify three damage grades (heavy,
extreme, destruction) in pre- and post-event DIM point clouds of an earthquake event in L’Aquila (Italy) in 2009. 
Through the identification of object-specific change features, our approach is transferable with respect to multi-source input point clouds used for training (virtual laser scanning) and application (real-world photogrammetry) of the model. We further achieve geographic transferability of the model by training it on simulated data of geometric change which characterises relevant damage grades across different geographic regions. As a result, the model yields high multi-target classification accuracies (overall accuracy:~92.0\%~-~95.1\% ; F1 score:~78.1\%~-~91.7\%). Its performance improves only slightly when using real-world region-specific training data ($<$~3\% higher overall accuracies;   $<$~3\% higher F1 scores) and
when using real-world region-specific training data ($<$~2\% higher overall accuracies; $<$~3\% higher F1~scores). Moreover, the detection rate for damaged buildings increases only slightly (+3.1\%) when using real-world instead of simulated point clouds as training data. We consider our approach especially relevant for applications where timely information on the
damage situation is required and sufficient real-world training data is not available.

\end{abstract}

\begin{keyword}
change detection \sep UAV \sep 3D \sep damage classification  \sep earthquake \sep natural hazards


\end{keyword}

\end{frontmatter}

\nolinenumbers


\section{Introduction}
\label{sec:introduction}
The timely assessment of earthquake-induced building damage after an earthquake event is of utmost importance for the effective planning of rescue and remediation actions. Automatic damage assessment based on the analysis of 3D point clouds (e.g., from photogrammetry or laser scanning) can provide fast and objective information on the damage situation within few hours \citep{ duarte_2020, vetrivel_2018}. As building damages can be of different type and degrees, a detailed assessment of multiple damage grades is required. This enables efficient use and distribution of resources, and supports the evaluation of structural stability of buildings and repair measures. 

The assessment of different damage grades, beyond binary damage detection, is a challenging task. There is large variety in possible damage characteristics and the transferability of methods developed for a study site to other geographic regions is limited, as is the transferability to data from other sources, especially for machine learning classifiers \citep{kerle_2020, nex_2019}. An approach for detailed assessment of structural building damage through 3D point cloud classification that is transferable both geographically and with respect to the source and characteristics of point clouds used for training and evaluation would strongly support damage assessment for earthquake response in urban areas.

The use of simulated point clouds for training a classifier might enable sufficient accuracies for damage classification. In case of an earthquake event, this can save valuable time as pre-trained classifiers can directly be applied to assess damage in event-specific real-world datasets without time-consuming manual labelling and further training using event-specific data.
In this paper, we therefore present a method of classifying damage grades in a supervised machine learning approach using a random forest classifier which is trained on simulated point cloud data. We assess how the use of generic simulated training data, instead of region-specific building and damage structures, influences the accuracy of identifying damage grades. The presented method contributes to a timely assessment of multi-class structural building damage in an earthquake event. The method can make use of simulated training data covering the full range of expected damage patterns to classify building damage in newly acquired point clouds using UAV-borne laser scanning or photogrammetry. If a high degree of transferability is achieved both geographically and with respect to the source of input point clouds used for training and application, the timely damage information provides great support for local rescue teams, as no additional building labeling is required after the occurence of a specific earthquake event.

\subsection{UAV-based point clouds for damage assessment}
New possibilities for structural building damage assessment have opened up in recent years with the increasing availability of UAV-borne remote sensing strategies. UAV-borne laser scanning~(ULS) or UAV-borne photogrammetry-based dense image matching~(DIM) provide 3D point clouds of urban quarters and entire cities within reasonable time frames (several hours). In disaster situations, the deployment of UAVs and even coordination of fleets is a crucial aspect to support local rescue teams with timely damage information \citep{meyer_2022}. 
Acquired point cloud data provides full-3D information of the captured scene and complements traditional image-based approaches \citep{stilla_2023, lin_2021, munawar_2021, galareta_2015}. Whereas images, i.e. photography, are suited to identify heavy roof damages and full building destruction, a more detailed assessment of damage grades requires a full-3D representation of the building geometry including, e.g., façades \citep{kharroubi_2022, Kohns_et_al_2021a, xu_2021}. Besides this geometric limitation, another challenge of image-based analysis is the strong variability of radiometric properties of objects over time or in different geographic regions \citep{qin_3d_2016}. Where topographic data is available, e.g. through 3D reconstruction using dense image matching, raster-based analysis is limited by the strong vertical component of building elements, which cannot be adequately represented from a 2D top-down view \citep{stilla_2023}. 

To obtain 3D point clouds, UAV-borne acquisition strategies allow a dense 3D reconstruction of the scene with point spacings down to a few centimetres. This allows change detection on the scale of individual building parts. These are important for identifying damage patterns that are typical for higher damage grades (heavy damage, extreme damage, destruction), which are target classes of our method (cf.~section~\ref{sec:methods}). The finer-scale geometry of small cracks or spallings can typically not be resolved in UAV-borne point clouds and their detection would require complementary higher-resolution acquisition. This can be achieved, e.g., using terrestrial laser scanning or close-range photogrammetry for selected buildings or blocks \citep{xu_2021}, but is not within the scope of our approach of building damage assessment. 

\subsection{Approaches for multi-class damage classification in 3D point clouds}
Structural building damage assessment based on UAV-borne point clouds so far has mainly been assessed using data acquired after an earthquake event, i.e.~using mono-temporal approaches. These data and approaches lack a-priori information in terms of pre-event data on the building structure. The information lack can be compensated with assumptions on the pre-event building shape, for example, to detect missing elements in the post-event point cloud \citep{vetrivel_2015}. This leads to misclassification where these assumptions do not hold true, and thereby limits the applicability and usability of mono-temporal approaches \citep{vetrivel_2016}.

Multi-temporal point cloud-based approaches so far have been constrained by the lack of pre-event datasets of earthquake-affected regions, which limited their practical applicability for real earthquake events. With the increasing availability of 3D city models and country-wide acquisitions of medium- to high-resolution 3D point clouds (e.g., through airborne laser scanning), the development of point cloud-based methods for damage asseessment through change detection between pre- and a post-event point clouds has increased in recent years \citep{deGelis_2021, xu_2021}. Current multi-temporal approaches directly compare a pre-event dataset and a post-event dataset and thereby extract different types of change features, e.g. geometric and radiometric change features. Change can also be extracted from machine learning approaches, such as random forest. Such supervised approaches classify building damage based on a set of geometric or histogram-based features (Roynard et al. 2016).

In general, multi-temporal approaches can be grouped into categories of post- and pre-classification. Post-classification approaches first apply a semantic segmentation of the dataset into different object types, and then analyse the target objects with respect to various kinds of changes between two epochs \citep{siddiqui_2017, awrangjeb_2015, xu_2015}. Pre-classification approaches first extract spatial areas of change in the point cloud and then classify these areas according to the extracted change \citep{xu_2015}. There also exist studies that combine both classification and change detection in one step \citep{tran_2018}.

For binary classification tasks, deep learning approaches today represent the state-of-the-art \citep{deGelis_2021, Ma_2019}. As such, they distinguish damaged and non-damaged buildings, or between two damage grades with very different damage characteristics \citep{kalantar_2020, nex_2019}. 
Existent methods of supervised machine learning classification are still challenged, though, by the variety of damage characteristics (mm crack widths and spalling for low damage grades up to partial failure modes and complete collapse for high damage grades), and by the transfer of trained algorithms to unseen data and other geographic regions \citep{huang_2019, vetrivel_2018}. 

\subsection{Training data generation with virtual laser scanning~(VLS)}
A prerequisite for machine learning-based damage classification is the availability of sufficient amounts of labelled training data covering the full range of damage patterns expected in an earthquake event \citep{deGelis_2021}. Lack of suitable training data for the classification task at hand can lead to poor classifier performances when transfered to an unseen dataset or geographic region \citep{munawar_2021, vetrivel_2018}. Training data demands of state-of-the-art machine learning approaches are difficult to meet when multiple damage grades shall be classified \citep{alzubaidi_2021}. In practical use, resulting inter-class confusion might lead to missing damaged buildings (i.e., classifying damaged buildings as undamaged), which are consequently not in the focus of local response teams. 
If no or insufficient labelled real-world data is available, training and evaluation of machine learning classifiers can benefit from simulated data \citep{deGelis_2021}. Whereas there are no tools available for the simulation of photogrammetric point clouds, multiple tools for the simulation of laser scanning point clouds exist \citep{winiwarter_2022, gastellu_2021, north_2010}.

Virtual laser scanning (VLS) is a powerful tool that provides simulated point clouds with known properties from labelled 3D input scenes \citep{hildebrand_2022}. Thereby, real-world scenarios of laser scanning acquisitions are recreated virtually \citep{winiwarter_2022}. Large amounts of realistic training data can hence be generated by VLS, which complement or replace real-world training data where the acquisition and labelling of sufficient amounts of real-world data is not feasible. By using VLS, training data can be automatically generated, and covers the full spectrum of relevant damage patterns. 
Even if yielding lower classification accuracies, training purely on simulated training data might provide adequate performance in time-sensitive situations, such as an earthquake event. Pre-trained classifiers can then be directly applied to assess damage in event-specific real-world datasets without further training.

Adequate modelling of damage patterns in the input scenes is thereby crucial for the accurate representation of damage in the simulated point clouds. This modelling process can, for example, be supported by domain-knowledge in earthquake engineering. A so-called damage catalogue, as developed by \cite{Kohns_et_al_2021a}, categorises typical geometric damage patterns for five different damage grades (slight, moderate, heavy, extreme, destruction) based on the European Macroseismic Scale~98 (EMS-98), following Grünthal (1998). It covers events in different geographic regions within Europe by considering observations of previous earthquake events and influences of material, building design, and structure on potential damage characteristics. Moreover, the damage catalogue by \cite{Kohns_et_al_2021a} has specifically been designed as decision basis for UAV-based assessment of structural building damage and therefore focuses on damage patterns of building elements that are recognisable from outside. Such domain knowledge allows to adequately represent damage patterns for the target damage grades in the simulated point clouds and ultimately, to discriminate multiple damage grades in classification tasks.

Real-world point clouds of an earthquake-affected area are more commonly derived from UAV-borne photogrammetry-based DIM than UAV-borne laser scanning, due to lower costs and wider availability of the instruments. However, photogrammetric point cloud simulation tools are not available for generating training data. Methods for damage assessment using simulated point clouds as training data therefore have to deal with different point cloud sources being used for training and application of the model. This can be possible if damage assessment is based on object-specific rather than data-specific features, i.e. features robust to different input point clouds.

\subsection{Objective}
\label{sec:objective}
\begin{figure}[]
    \centering
    \includegraphics[scale=1.2]{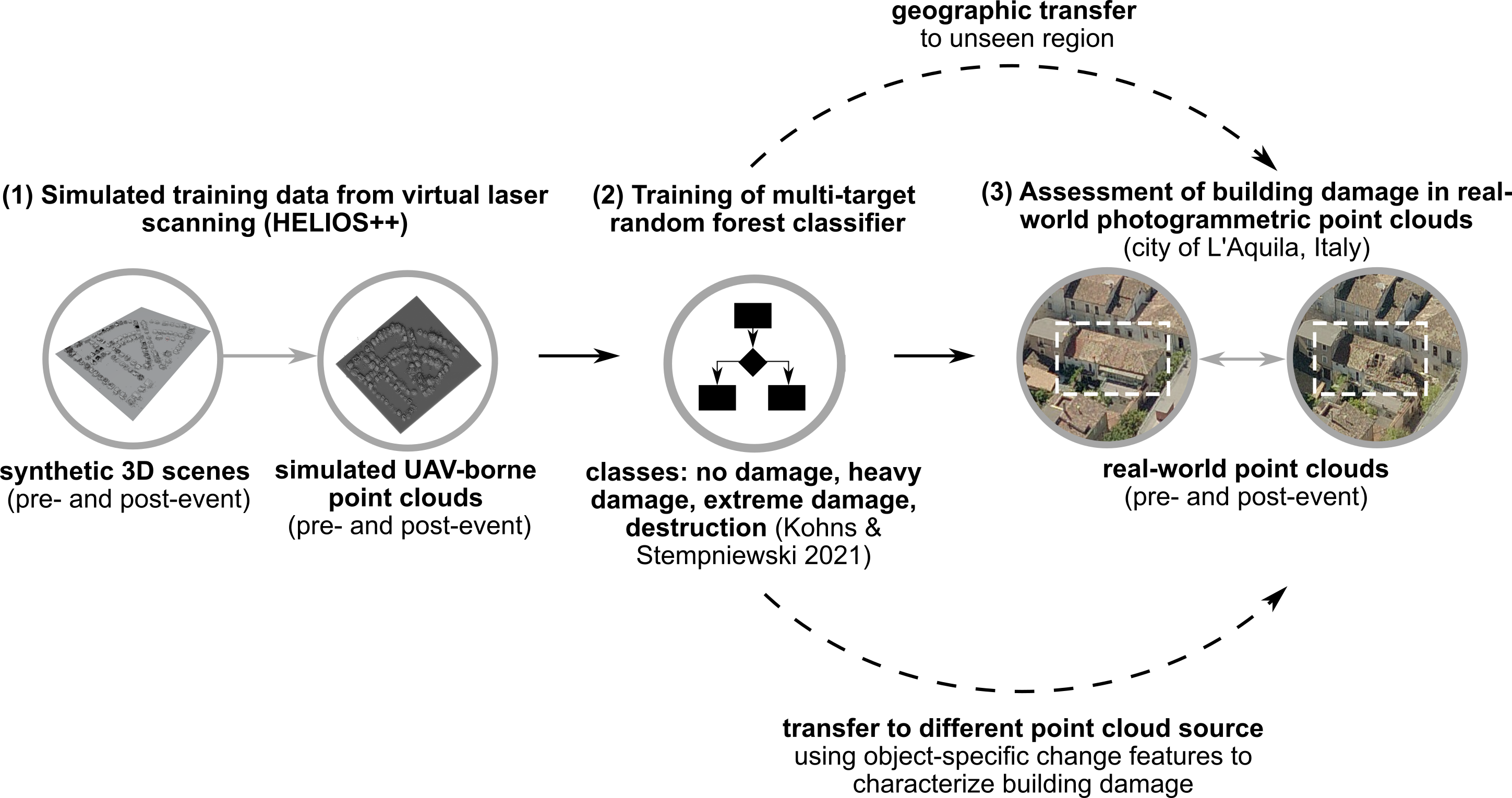}
    \caption{Overview of the approach to classify building damage in real-world photogrammetric point clouds using a machine learning model trained on simulated laser scanning point clouds.}
    \label{fig:graphical_abstract}
\end{figure}

In this research, we automatically classify multi-class structural building damage from multi-temporal real-world photogrammetric point clouds using a random forest model trained on virtual laser scanning data (Figure~\ref{fig:graphical_abstract}). We develop our method to consider the following aspects:
\begin{enumerate}
 \item Damages are assessed per building by deriving change of geometric features between pre-event and post-event point clouds. We are thereby independent from modelling of pre-event building shapes, but can derive change through the comparison of multi-temporal point clouds. 
 \item Domain knowledge is integrated from earthquake engineering in the process of training data generation from virtual scenes. Using a descriptive damage catalogue \citep{Kohns_et_al_2021a} we include knowledge on possible damage patterns for different damage grades and consider regional and building-specific variability. We thereby ensure that our training data covers the full spectrum of damage patterns expected in the real-world dataset. 
 \item By using virtual laser scanning~(VLS) for the generation of simulated point clouds, labelled building-specific training data with realistic point cloud characteristics is automatically obtained.
 \item Through the use of object-specific change features, our machine learning model is trained on simulated UAV-borne laser scanning point clouds to classify damage in real-world UAV-borne point clouds derived from photogrammetry-based dense image matching (DIM). The aim is to achieve transferability with respect to the source of input point clouds used for training and application of the model.
\end{enumerate}

\newpage

With our developed approach, we complement existing methods for multi-class structural building damage assessment especially for applications where timely damage information is required and suitable, and sufficient pre- and post-event real-world training data is not available. To possibly increase timeliness of response further, we investigate how the use of non-region-specific training data influences the classification accuracy to thereby assessing transferability between geographic regions.

\section{Study Site and Data}
\label{sec:study_site_data}
In this paper, we use UAV-borne DIM point clouds of the city of L'Aquila, Italy, to classify structural building damage with our method. 
L'Aquila was hit by an earthquake on Monday, April 6, 2009 at 3:32 a.m.~local time in the central Italy region of Abruzzo with a moment magnitude Mw~=~6.3 \citep{eer_2009, geer_2009}. The affected area is located within the central section of the Apennines. This mountain chain results from the convergence between the African tectonic plate, to which it belongs, and the European tectonic plate, and the subsequent collision of the two continental margins \citep{geer_2009}. The epicenter was close to L’Aquila, a city with an approximate population of 73,000 inhabitants. As the earthquake occurred when most people were sleeping, over 300 people died and 1,500 were injured \citep{eer_2009}. This event was the strongest of a sequence that started a few months earlier. 56 of the approximately 300 digital strong-motion stations of the Italian Strong Motion Network recorded the main shock, whereby five stations were located within a 10 km radius around the epicenter. The latter all recorded horizontal peak ground accelerations above 0.35~g, and some showed permanent displacements up to 15~cm. The ground motion had a short duration of 10~s or less \citep{eer_2009}. Up to 15,000~buildings were destroyed or damaged, 70,000~to~80,000 people were temporarily evacuated, and more than 24,000 were left homeless  \citep{eer_2009}. Damage occurred in the city of L’Aquila, but more widespread extreme damage was seen in smaller towns such as Onna, Paganica, and Castelnuovo, where more than 50\% of the historic centers were damaged beyond repair. Collapsed and damaged structures in L’Aquila included both masonry buildings and reinforced concrete structures. Many relatively modern reinforced concrete frame structures with masonry infills were not damaged. Where damage occurred, it generally involved minor cracking to relatively severe cracking or collapse of the masonry infill walls. If there was heavy damage in older reinforced concrete buildings, it most likely resulted from the lack of ductility, and the brittleness of exterior infill walls and interior partition walls  \citep{geer_2009}. Old unreinforced masonry buildings made of mortar and multi-wythe rubble-stone or clay bricks were significantly damaged, ranging from wall cracking to extreme damage and collapse. The damage indicated strong effects of site conditions. High damage levels were seen in villages built at least partly on relatively young sediments, but only slight damage was visible in neighbouring villages on solid rock materials, for buildings with similar material quality and characteristics  \citep{eer_2009}.

DIM point clouds of the city of L'Aquila were generated based on oblique and nadir RGB images captured before (2008-08-30) and after (2009-04-29) the earthquake event. Point clouds were generated based on these images through dense image matching using Agisoft Metashape~(version 1.8.2). 
The resulting point clouds contain around 72~million and 78~million points with an average point spacing of 0.1~m.
We improved the alignment of the pre- and post-event point clouds using an iterative closest point algorithm \citep{besl_mckay_1992} applied to stable areas (streets and stable walls) in the point cloud. We assess the accuracy of the alignment between the point clouds by calculating the standard deviation of M3C2 distances \citep{lague_accurate_2013} in stable areas, following \cite{zahs_2022}.
As a result of the nadir perspective of the UAV images acquired to generate this point cloud, horizontal building elements (roofs) were sampled with a higher density (point spacing: 0.09~m; STD.:~0.03~m) compared to vertical elements (point spacing:~0.12; STD.:~0.04~m).

As additional data, we use simulated point clouds from virtual laser scanning as training data for our random forest classifier. The generation of the simulated dataset is described in the methods (section~\ref{sec:methods}).

\textbf{\begin{figure}[]
    \centering
    \includegraphics[]{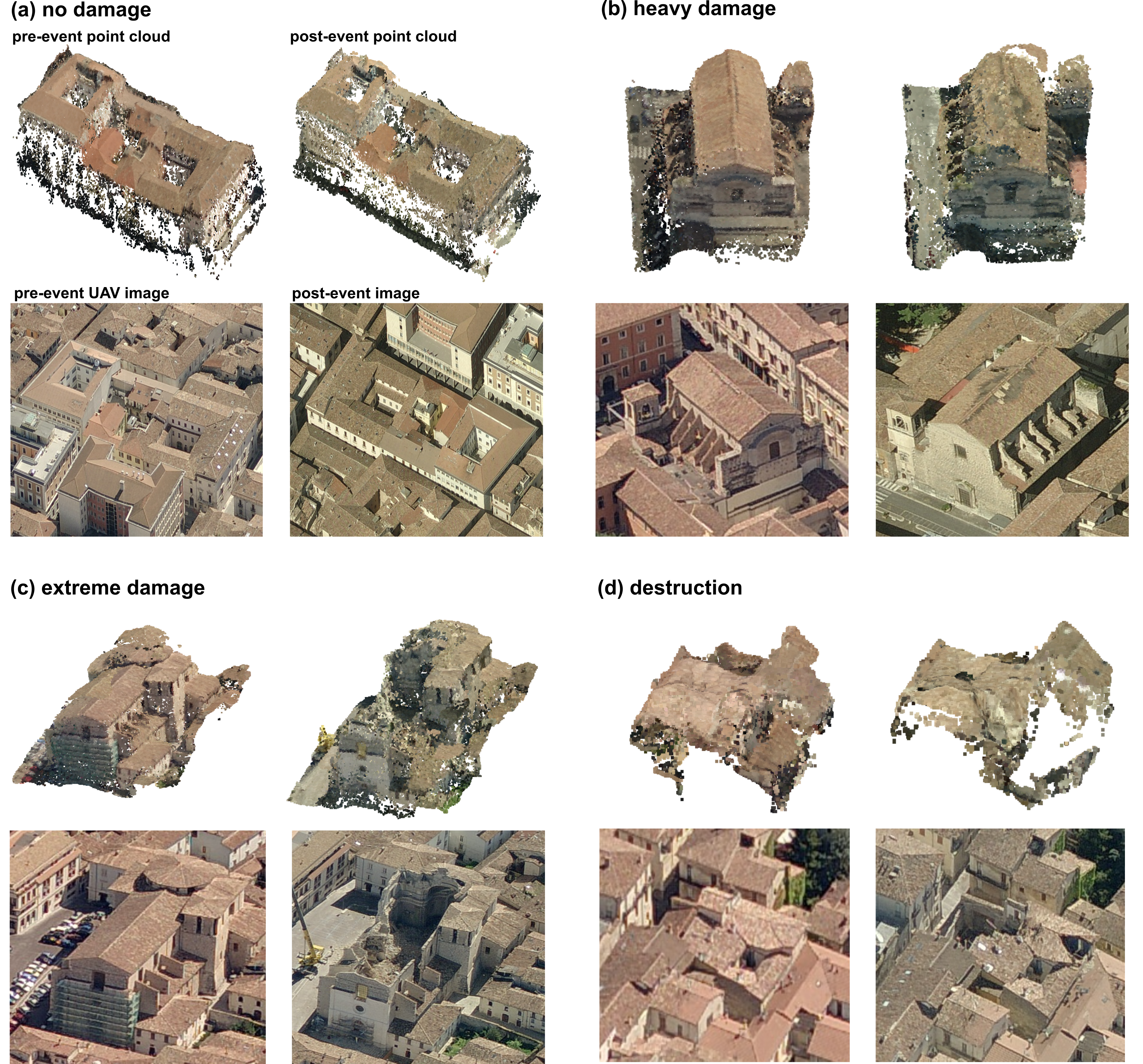}
    \caption{Pre-event and post-event building point clouds and corresponding UAV images representing the four target damage grades (a) no damage, (b) heavy damage, (c) extreme damage, and (d)~destruction.}
    \label{fig:damages_laquila}
\end{figure}}

\clearpage

\section{Methods}
\label{sec:methods}
\begin{figure}[]
    \centering
    \includegraphics[]{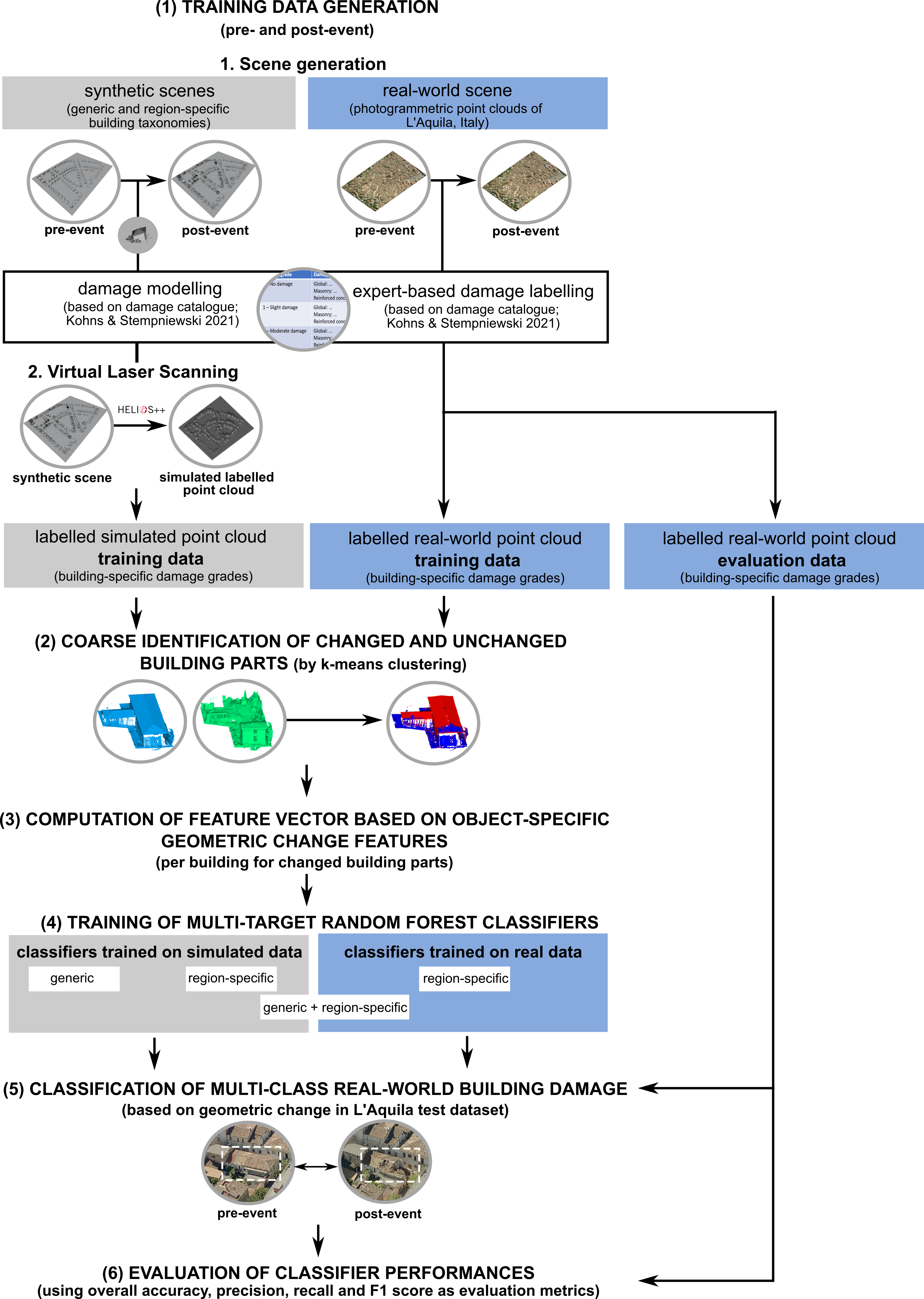}
    \caption{Full workflow of the approach including (1)~the generation of simulated and real-world training data, (2)~coarse identification of changed and unchanged building parts, (3)~computation of object-specific geometric change features, (4)~training of multi-target random forest classifiers, (5)~classification of multi-class building damage in a real-world UAV-borne photogrammetric dataset, and (6)~evaluation of classifier performances.}
    \label{fig:Figure1}
\end{figure}

In our study we classify multiple grades of structural building damage in a real-world UAV-borne DIM point cloud using a machine learning model trained on virtual UAV-borne laser scanning point clouds (Fig.~\ref{fig:Figure1}). 
Following domain knowledge in earthquake engineering (cf. section~\ref{sec:methods_scenes}) we consider four damage grades in our approach (Fig.~\ref{fig:Figure2}): No damage, heavy damage, extreme damage, destruction. Classes of slight and moderate damage are not considered, as the geometric representation of their typical damage patterns (e.g., crack widths of few millimetres) in the point clouds requires a higher spatial data resolution, i.e. more detailed acquisitions than typically available by UAV acquisitions.
Our approach consists of five main steps:
\begin{itemize}
\item Simulation of pre- and post-event point clouds through virtual UAV-borne laser scanning of virtual scenes
\item Coarse identification of changed and unchanged building parts using a k-means clustering approach
 \item Extraction of robust object-specific change features
 \item Training of a random forest machine learning model with simulated point clouds and object-specific change features
 \item Random forest classification of mulit-class building damage in real-world pre- and post-event DIM point clouds
\end{itemize}

\textbf{\begin{figure}[]
    \centering
    \includegraphics[]{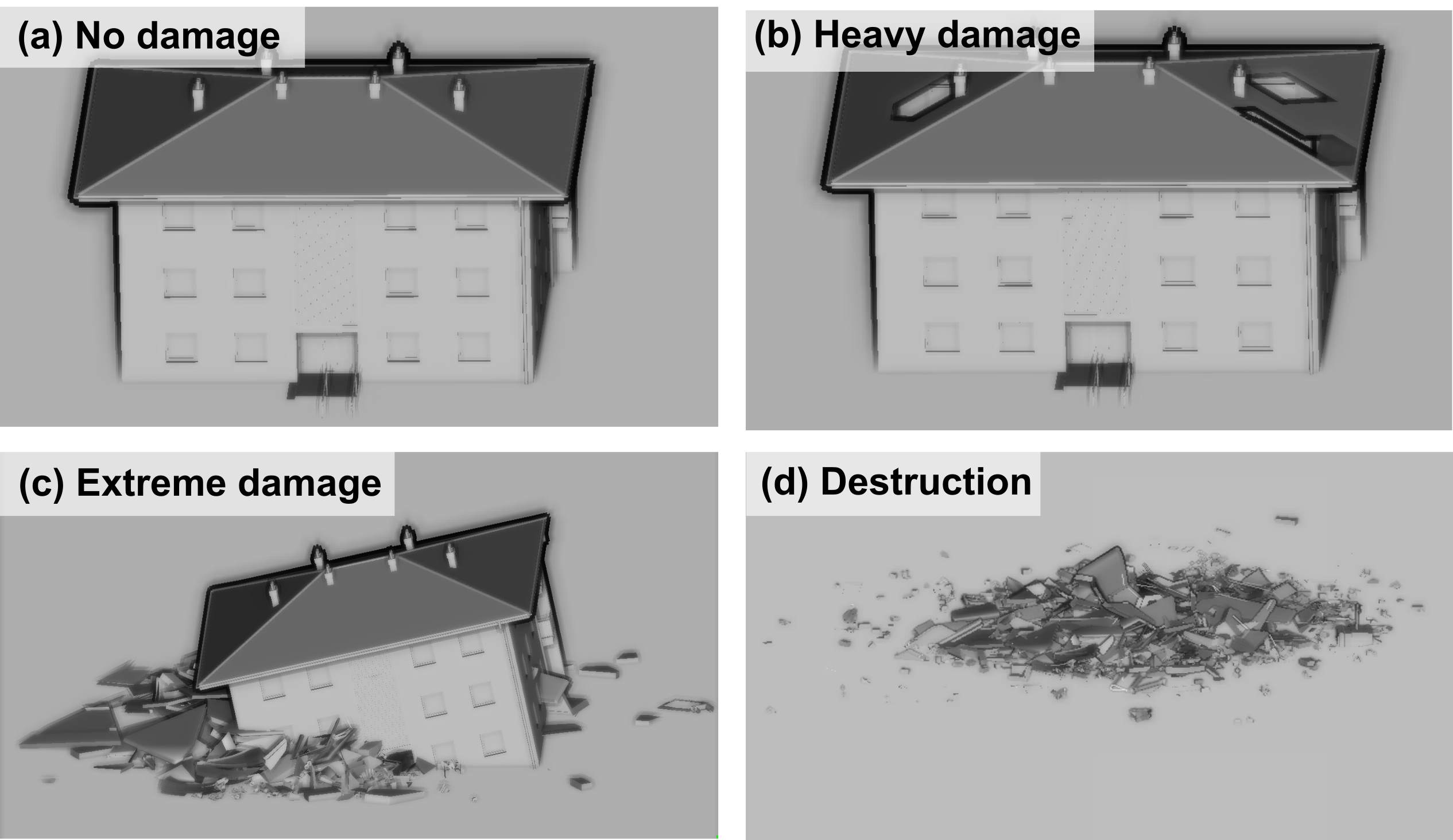}
    \caption{Example 3D building model representing the four target damage grades (a)~no damage, (b)~heavy damage, (c)~extreme damage, and (d)~destruction considered in our classification.}
    \label{fig:Figure2}
\end{figure}}

We evaluate the performance of classifiers trained on (1)~simulated generic ULS point clouds, (2)~simulated region-specific ULS point clouds, (3)~simulated generic ULS point clouds and real-world region-specific DIM point clouds, and (4)~real-world region-specific DIM point clouds with respect to their capability to accurately classify building damage in a real-world DIM dataset. Evaluation is based on a reference dataset derived from the real-world DIM point clouds of the L'Aquila earthquake in 2009 (cf.~section~\ref{sec:study_site_data}) with the help of the engineering expert. A set of evaluation metrics (overall accuracy, precision, recall, F1 score) is used to assess the performance for each target damage grade separately as a binary case and the overall capacity of the classifiers to correctly separate
buildings of any damage grade from undamaged buildings.

\subsection{Generation of real-world training and evaluation data}
\label{sec:methods-DIM}
The DIM dataset of L'Aquila is used to generate both real-world training data and to evaluate the performance of all classifiers with respect to their ability to assess structural building damage in real-world DIM point clouds. 
We introduce a proper split of the real-world dataset into a training dataset and an evaluation dataset, and use the training dataset exclusively for the generation of real-world training data. 
Areas of individual buildings in the dataset are manually segmented and labelled by an expert in earthquake engineering, and by using the damage catalogue developed in \cite{Kohns_et_al_2021a} to identify damage patterns typical for the target damage grades. 
The training dataset consists of 112 labelled building models per damage grade. The evaluation dataset consists of 125~labelled buildings in total (35~no damage, 19~heavy damage, 32~extreme damage, 35~destruction). The uneven distribution of target damage grades in the evaluation dataset results from the uneven number of buildings per damage grade that could be confidently assessed in the manual expert-based labelling. 

For each damage grade considered in this study, Figure~\ref{fig:damages_laquila} shows exemplary building point clouds extracted from the real-world dataset of L'Aquila for the purpose of training the machine learning model. Damage patterns in the training samples cover all patterns typical for each of the damage grades and include (1)~total collapse (destruction), (2)~partial collapse of stories, fa\c{c}ades and roofs (extreme damage), (3)~large holes distributed across all building elements (heavy damage), and (4)~no damage.
Buildings used for training within the entire real-world dataset are excluded from the evaluation dataset.

Some buildings in the evaluation dataset could not be labelled due to difficulties or uncertainties in the recognition of damage patterns by the expert. This is mostly related to occlusion of large parts of the building due to vegetation or other scene objects.

\subsection{Generation of simulated training data}
\label{sec:methods-VLS}
The generation of simulated training data consists of two steps:
\begin{enumerate}
 \item Preparation of virtual scenes of 3D building models with various damage patterns, labelled with the damage grade
 \item Simulation of point clouds through virtual laser scanning of these virtual scenes
\end{enumerate}

\subsubsection{Preparation of virtual scenes}
\label{sec:methods_scenes}
3D building models used to assemble the virtual scenes are taken from open source online repositories for 3D models \citep{free3d_2023}. Further buildings are generated manually in the free and open-source 3D creation suite software blender \citep[version 2.93.0]{blender_2018}. The number of the originally 28 different building models is augmented by applying modification to building size or parts of the buildings. This results in a total number of 448 buildings composed of 112 buildings per damage grade.
To investigate the importance of region-specific training data, two types of virtual scenes in pre-event and post-event state are generated:
\begin{enumerate}
 \item Region-specific scene: This scene mimics the characteristics of the real-world scene of this study (L'Aquila) with respect to building types and construction materials typical for this region (cf.~section~\ref{sec:study_site_data}). It also exhibits main characteristics of the urban structure when assembling the individual 3D building models in the scene. Damage patterns implemented in the post-event point cloud are typical damage patterns for this geographic location (cf.~section~\ref{sec:study_site_data}). 
 \item Generic scene: This scene contains a broader range of building types (single family houses up to large apartment buildings), construction material (masonry and reinforced concrete), and built structure (both loose and narrow development), all of which typically occur in small to medium-sized European cities. Consequently, the post-event state of this scene contains a greater variety of damage patterns.
\end{enumerate}

Both types of scenes consist of 112 undamaged individual buildings, respectively.
The three post-event damaged scenes (heavy damage, extreme damage, destruction) are generated based on their corresponding pre-event scenes. Therein, we introduce damage representative of the respective damage grade to each building in duplicates of each pre-event scene. This provides the data basis for a direct comparison of pre-event and post-event scenes to extract change features and then to classify structural building damage in a later step.

Structural building damage is modelled into the buildings based on the damage catalogue developed by \cite{Kohns_et_al_2021a}. 
It defines distinct geometric properties typical for our target damage grades, which we use to discriminate them clearly in the classification.
For each damage grade considered in our study (no damage, heavy, extreme, destruction) we manually introduce the typical damage patterns into 3D building models of the virtual post-event scenes. We ensure the realism of introduced damage patterns and the correct labelling of each building with respect to its damage grade in a visual evaluation by the earthquake engineering expert.

Following the damage catalogue \citep{kohns_2022, Kohns_et_al_2021a}, we focus on the two pre-dominant construction materials in Europe, i.e. masonry and reinforced concrete. We do not consider further building  material, such as wood or steel for the following reasons: Wood is flexible and able to dissipate lots of energy through the timber joints. Steel has a ductile material behaviour and only the connection points, which are difficult to assess from outside, are relevant in case of an earthquake event. In contrast to wood and steel, masonry and  reinforced concrete show distinct differences between damage grades. Related damage patterns are visible from the outside, which renders UAV-based damage assessment of entire quaters possible.

\subsubsection{Simulated point cloud generation using virtual laser scanning~(VLS)}
\label{sec:methods-vls}
We perform virtual laser scanning of the generated scenes using the open-source tool HELIOS++ \citep{winiwarter_2022}. 
HELIOS++ is a general-purpose ray tracing-based simulation framework with support for multiple platforms, sensors, and scene types that can be flexibly combined in a modular manner. 

Acquisition parameters (Tab.~\ref{tab:datasets_overview}) for our simulations are selected in accordance with point cloud characteristics of the real-world dataset (cf.~section~\ref{sec:methods-DIM}). Our goal is to achieve point densities similar to the real-world data in all simulated point clouds. The influence of different acquisition parameters between pre- and post-event acquisitions on the the geometric representation of a building and consequently on the classification performance is, for example, assessed by \citep{deGelis_2021} and is not in the focus of our study. 

\begin{table}[]
\small
\centering
\caption{Acquisition parameters used for the simulation of UAV-borne laser scanning in HELIOS++.}
\vspace{12pt}
\begin{tabular}{ccccccc}
 \textbf{\begin{tabular}[c]{@{}c@{}}Scan rate [lines/s]\end{tabular}} & \textbf{\begin{tabular}[c]{@{}c@{}} Pulse \\ repetition rate \\ {[}kHz{]}\end{tabular}} 
 & \textbf{\begin{tabular}[c]{@{}c@{}}Strip overlap \\ {[}\%{]}\end{tabular}} 
 & \textbf{\begin{tabular}[c]{@{}c@{}}Field of \\view \\ {[}deg.{]}\end{tabular}} & \textbf{\begin{tabular}[c]{@{}c@{}}Flight altitude \\ {[}m AGL{]}\end{tabular}}                                         & \textbf{\begin{tabular}[c]{@{}c@{}}Flight speed \\ {[}m/s{]}\end{tabular}} \\ \hline  89                                                                  & 300   & 60                                                                        & 120                                      & 100                                                           & 8                                                                                                                   
\end{tabular}
\label{tab:datasets_overview}
\end{table}

As input for the simulations we specify one virtual scene, respectively. The damage label and unique ID annotated to each building are passed through the simulation process and stored with the output VLS point cloud.
As output of the simulation we obtain pre- and post-event region-specific and generic ULS point clouds with per-building damage grades as class label in the post-event point clouds.
We segment individual building point clouds based on the building ID of the virtual scene stored with each point in the output point cloud of the simulation.

\subsection{Object-specific change feature selection}
\label{sec:methods-features}
We assess structural building damage through change analysis of geometric features between point clouds of pre- and post-event epochs. As we evaluate the use of different sources of input point clouds for training (VLS) and for classification (real-world DIM), we investigate the transfer of change features to different sources of input point clouds in an experimental study of real-world ULS and real-world DIM point clouds. 
We use the experiment to identify object-specific change features that are robust to variable properties of input point clouds from different sources (laser scanning and photogrammetry). 
We consider a geometric change feature (e.g., change in curvature between two epochs) as robust if the relative difference of the change of this feature between two epochs is low between laser scanning and DIM point clouds. The absolute value of a geometric feature in one epoch can differ to a larger extent between a ULS and a DIM point cloud as different close-range remote sensing techniques provide data with different inherent properties and represent objects in a different way \citep{mandlburger_2017}.

Object-specific change features are identified in point clouds of building damages acquired via ULS and DIM, which allows us to directly assess the robustness of geometric features to real-world ULS and DIM point clouds as input.
We use real-world point clouds acquired of a building at three epochs during a demolition process. Pre-event ULS and DIM point clouds were captured before the demolition, and post-event 1 and 2 ULS and DIM point clouds were captured at two stages of the demolition process (Figure~\ref{fig:inf288}). To make results of this experiment applicable to real-world and simulated datasets used in this study, we subsample the demolition point clouds to match the point density of the L'Aquila dataset and simulated point clouds (0.1~m point spacing, STD:~0.04~m). Similarly, point densities are higher on the roofs (96~pts./m2) than on the fa\c{c}ades (66~pts./m2).

Geometric features are computed per point with a certain local neighbourhood radius. Change of a feature between pre- and post-event epochs is then obtained by computing the difference between the feature value of a point in the pre-event epoch to the feature value of it's closest point in the post-event epoch using a kd-tree-based nearest neighbour search.
We investigate the following hand-crafted features which are commonly used in classical machine learning approaches of building damage assessment \citep{deGelis_2021, tran_2018}: Linearity, planarity, omnivariance, surface variation, curvature, point density, no. neighbours, surface density, volume density, sphericity, verticality, eigentropy, anisotropy, eigenvalues, sum of eigenvalues, roughness, z rank (relative position of the feature point within its vertical neighbourhood), z range (highest minus lowest z value), normal vector, and echo ratio \citep{hoefle_2009}.

These features are computed for local neighbourhood radii in a range of 1.0~m to 4.0~m at a step width of 0.5~m, according to the point density of the dataset. We finally select the neighbourhood size where relative change of the features is most similar between laser scanning and photogrammetry-based analyses. 
Therein, we consider only those geometric features as input for damage classification which show up to 10\% difference in relative change between ULS and DIM point clouds of two epochs. The threshold of $\leq$~10\% is selected as the number of features is most stable between 10\% and 15\% difference (Figure~\ref{fig:pareto}).
\textbf{\begin{figure}[]
    \centering
    \includegraphics[]{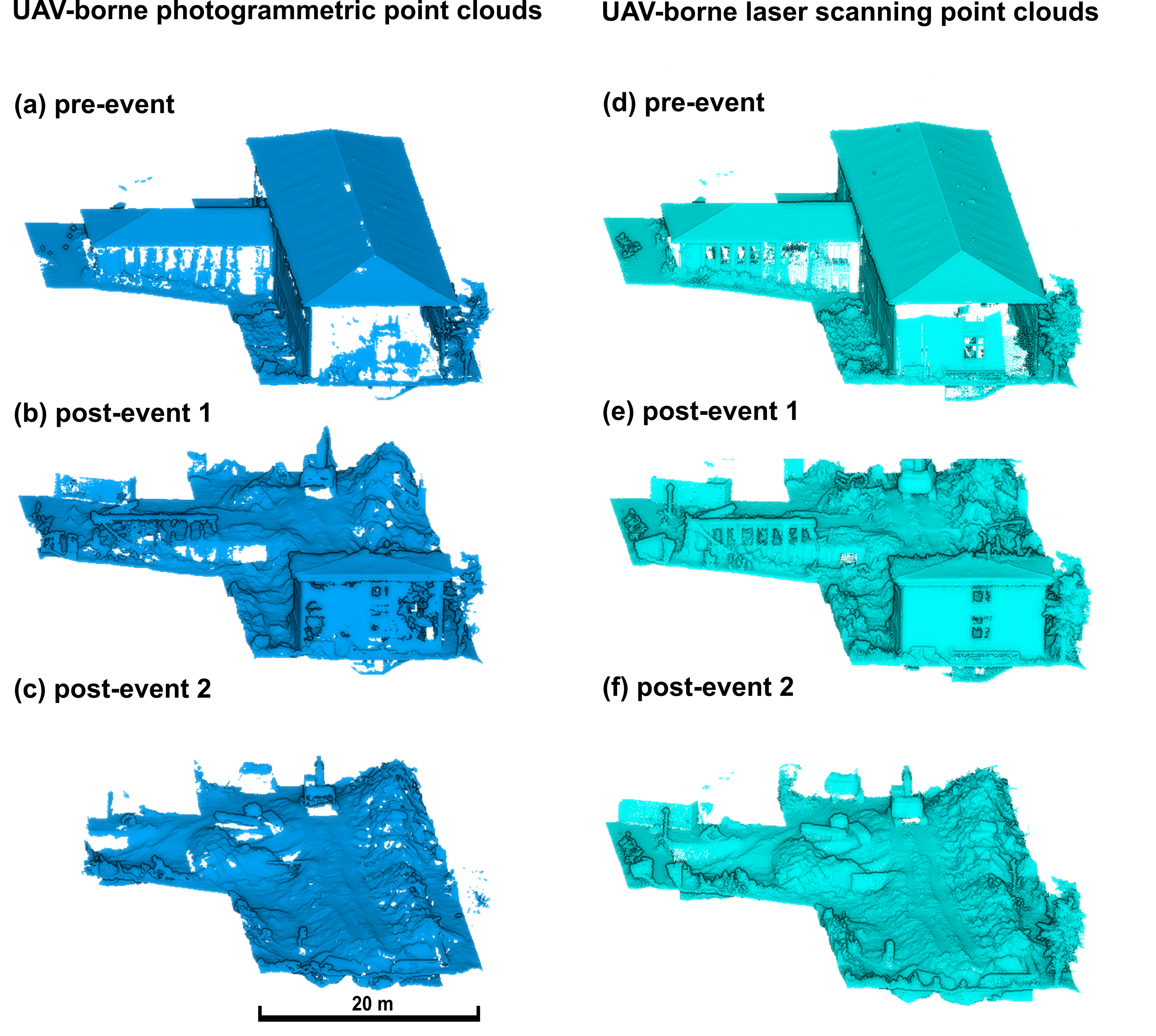}
    \caption{(a)-(c) Real-world UAV-borne photogrammetric point clouds and (d)-(f) real-world UAV-borne laser scanning point clouds acquired of a building before demolition (a-d) and at two different demolition stages (b-f).}
    \label{fig:inf288}
\end{figure}}

\textbf{\begin{figure}[]
    \centering
    \includegraphics[]{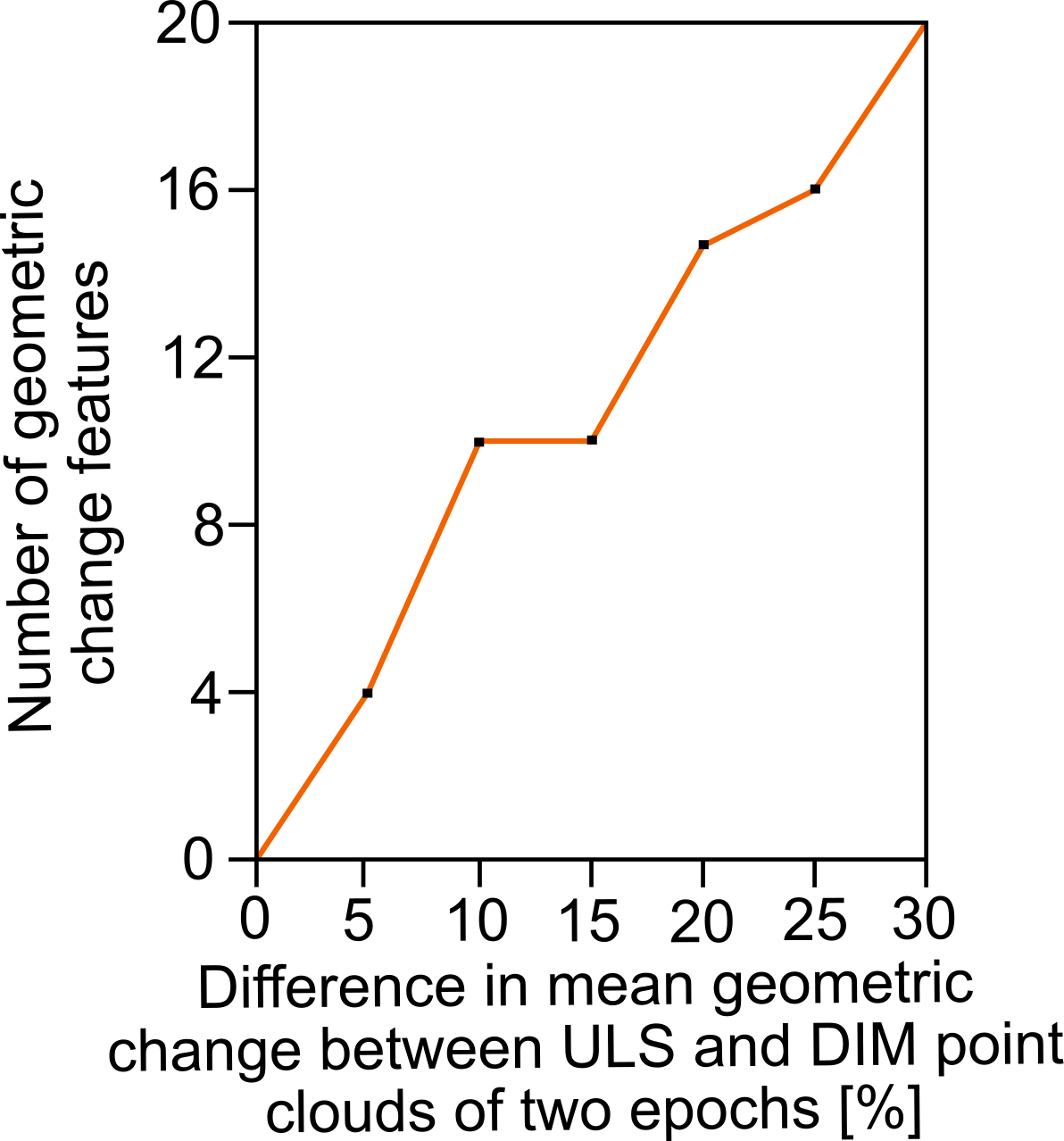}
    \caption{Number of geometric change features depending on the threshold used as allowed difference in mean geometric change between point cloud of UAV-borne laser scanning (ULS) and UAV-borne photogrammetry-based dense image matching (DIM) point clouds of two epochs in the experimental study.}
    \label{fig:pareto}
\end{figure}}
\subsection{Coarse extraction of changed building parts}
\label{sec:clustering}

Before the actual classification of damage grades, we coarsely filter out non-damaged parts using a k-means clustering on each building point cloud. 
This is done because even for heavy or extreme damage, larger parts of the building point cloud can still be unchanged. Hence, geometric change following typical damage patterns occurs only in local parts of a building. Descriptive statistical values (e.g.,~mean or median change of a geometric feature) per building are then not suitable to represent the actual degree of damage, as exemplary shown for 
a building of heavy damage in Figure~\ref{fig:clustering_building}.
A supervised machine learning model will consequently fail to correctly classify such building damage.

\textbf{\begin{figure}[]
    \centering
    \includegraphics[]{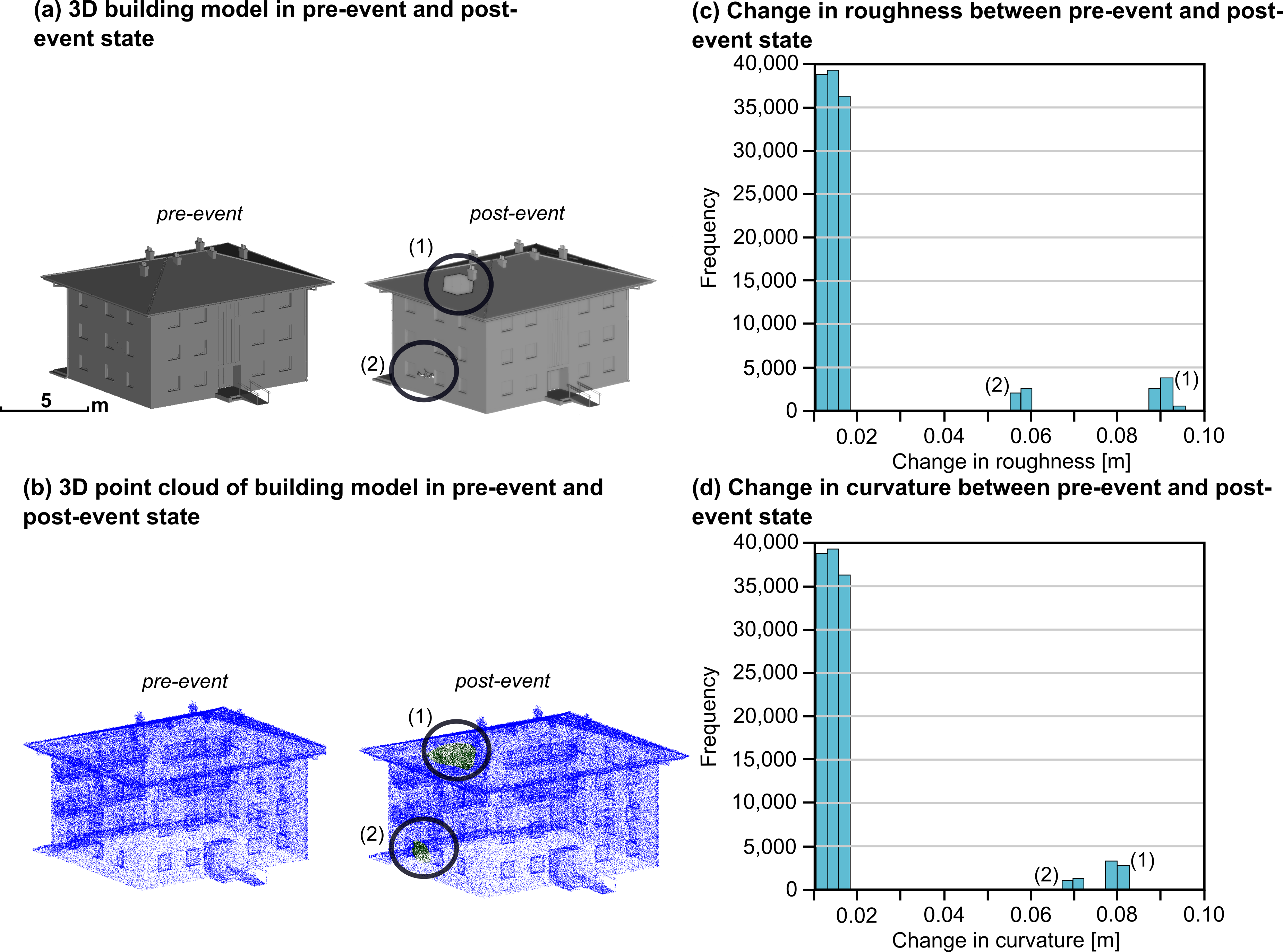}
    \caption{(a)-(b) 3D building model and derived point cloud of a building in a pre-event and post-event state. (c)-(d) histogram of change in roughness and curvature between pre-event and post-event state. Small building parts exhibiting change (1~and~2) result in a multi-modal distribution of change values, which cannot be adequately described by descriptive statistical values derived for the whole building.}
    \label{fig:clustering_building}
\end{figure}}
With our target damage grades, we can easily separate changed and unchanged building points in feature space using change in curvature (to identify holes and large cracks) and change in height (to identify collapsed roofs and stories).
We use clustering to separate all building points into two clusters (changed~/~unchanged), from which we only use points of the changed cluster for the classification of damage grades. Buildings with no changed points would directly be considered as undamaged buildings. 

To reflect the share of damaged area of a building after filtering, we include the share of damaged points as additional feature in the classification, derived as the percentage of clustered changed points to all points of a building.

\subsection{Classification of building damage grades}
\label{sec:methods-a´clf}
To classify structural building damage, we use supervised classification using the object-specific geometric change features for the changed points per building.
We use random forest decision trees for classification, which are suitable for our application as they are mostly uncorrelated due to high variations of the trees and do not require large amounts of training data \citep{breiman_random_2001}.

To investigate the influence of using region-specific and real-world training data for damage classification of structural building damage in real-world point clouds, we train multiple random forest classifiers with different input training data (cf. sections~\ref{sec:methods-DIM} and \ref{sec:methods-VLS}):
1)~Simulated generic VLS data (VLS~generic), 
2)~Simulated region-specific data (VLS~region specific),
4)~Simulated generic ULS data and real-world region-specific DIM data (VLS~generic~+~real-world DIM).
4)~Real-world region-specific DIM data (real-world~DIM).

Each classifier is trained and tested using the 448 labelled damaged and undamaged buildings of the respective dataset with an equal number of 112 buildings per damage grade. In the selection of the buildings used for training, we ensure that the full range of damage patterns typical for the respective damage grade are included in the training dataset. For each classifier, building objects of the entire training dataset are randomly split into 70\% training data and 30\% testing data to evaluate the accuracy of the trained classifier. The random forest classifier is trained with a set of 100 trees and a maximum depth of 5, which provides adequate capacity for this classification task. 
All four trained classifiers are finally applied to the real-world DIM dataset.

\subsection{Evaluation of classifier performances}
\label{sec:methods-clf_eval}
The performance of the classifiers with respect to their capability of accurately classifying structural building damage in a real-world DIM point cloud is assessed using the labelled evaluation dataset. The major focus of the evaluation is on (1)~the transferability with respect to multi-source input point clouds used for training and application of the model, and (2)~the geographic transferability of models trained on generic simulated data for classification of a real-world dataset.

We evaluate the performance for each target damage grade separately as a binary case, as the correct discrimination of multiple damage grades is of great relevance to our study. 
Further, we evaluate the overall capacity of the classifiers to correctly separate buildings of any degree of damage from undamaged buildings.
We use overall accuracy, precision, recall, and F1 score as classification metrics to assess the classifiers performances.
\clearpage

\section{Results}
\label{sec:results}
\subsection{Generation of simulated training data}
\label{sec:simulated_training_data}
The two types of virtual scenes (generic and region-specific) are shown in Figure~\ref{fig:virtual_scenes} in pre-event and post-event states along with exemplary damage patterns of various 3D building models with different damage grades. All buildings of the pre-event scene are modified to represent damaged buildings of the damage grades in the post-event states. 
\textbf{\begin{figure}[]
    \centering
    \includegraphics[]{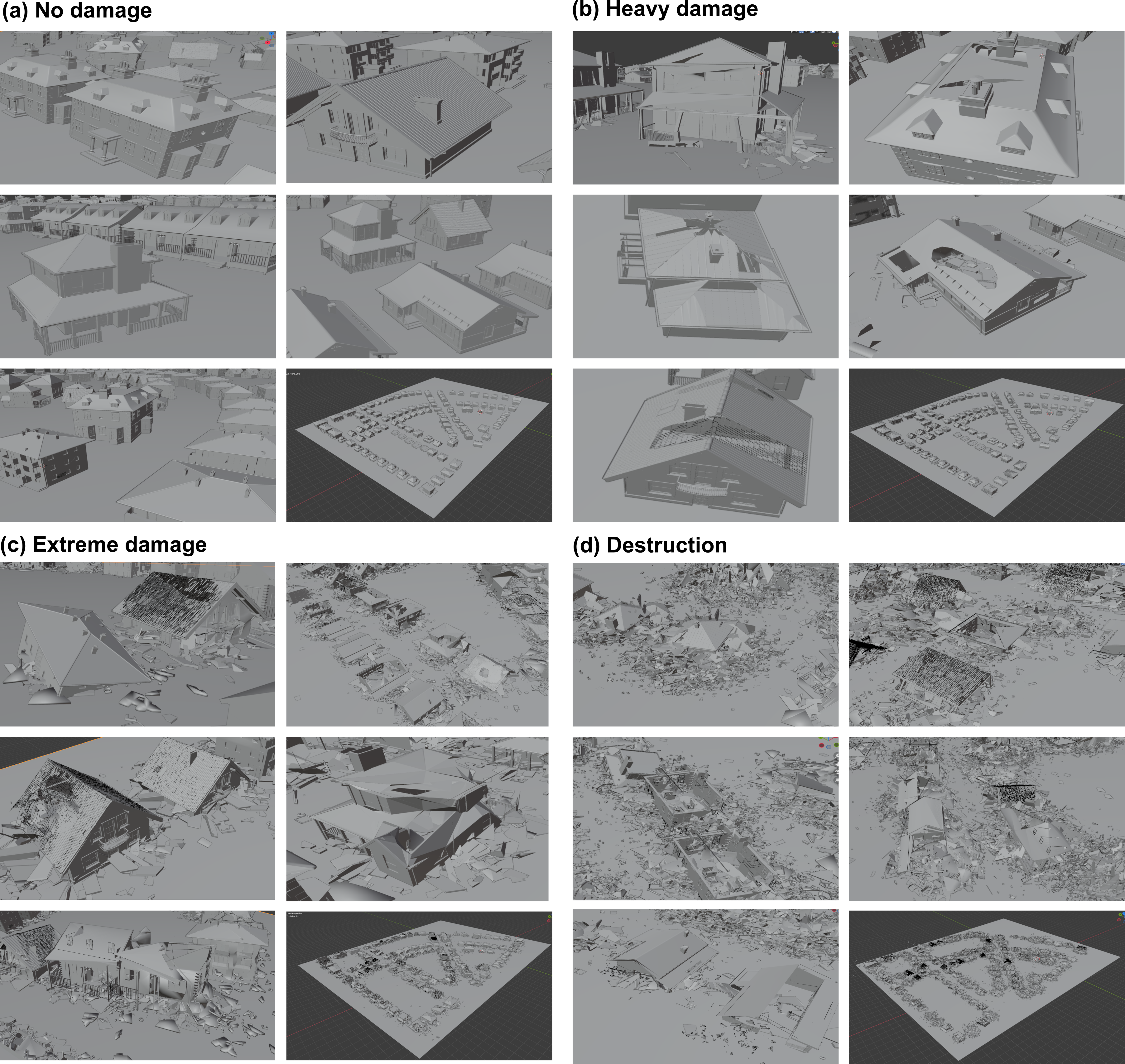}
    \caption{Examples of 3D building models of a generic scene in (a)~pre-event state and (b)~-~(d) post-event states of the target damage grades (a)~no damage, (b)~heavy damage, (c)~extreme damage, and (d)~destruction.}
    \label{fig:virtual_scenes}
\end{figure}}
Point clouds of the region-specific and generic virtual scenes are shown in Figure~\ref{fig:simulated_vs_realworld} as zoom-ins to individual buildings. As a result of the downward-looking perspective of the UAV-borne acquisition, buildings feature higher point densities on their horizontal elements (roofs; mean point spacing: 0.07~m, STD. 0.03~m) compared to their vertical elements (fa\c{c}ades; mean point spacing: 0.11~m , STD: 0.04~m). 
Figure~\ref{fig:simulated_vs_realworld} also compares real-world and simulated building point clouds of buildings with similar damage patterns to demonstrate similarities and differences in the geometric representation of damage. It is clearly visible that the spatial sampling of the buildings differs due to the different acquisition strategies and consequently different point cloud characteristics. The geometric representation of damage patterns is, however, not considerably affected by these differences, as for higher damage grades the change in geometry occurs on a larger spatial scale than the differences in sampling due to the overall dense sampling of a building. We can therefore expect that extracted geometric change between two epochs is in the same order of magnitude in both simulated laser scanning and real-world DIM point clouds.

\textbf{\begin{figure}[]
    \centering
    \includegraphics[]{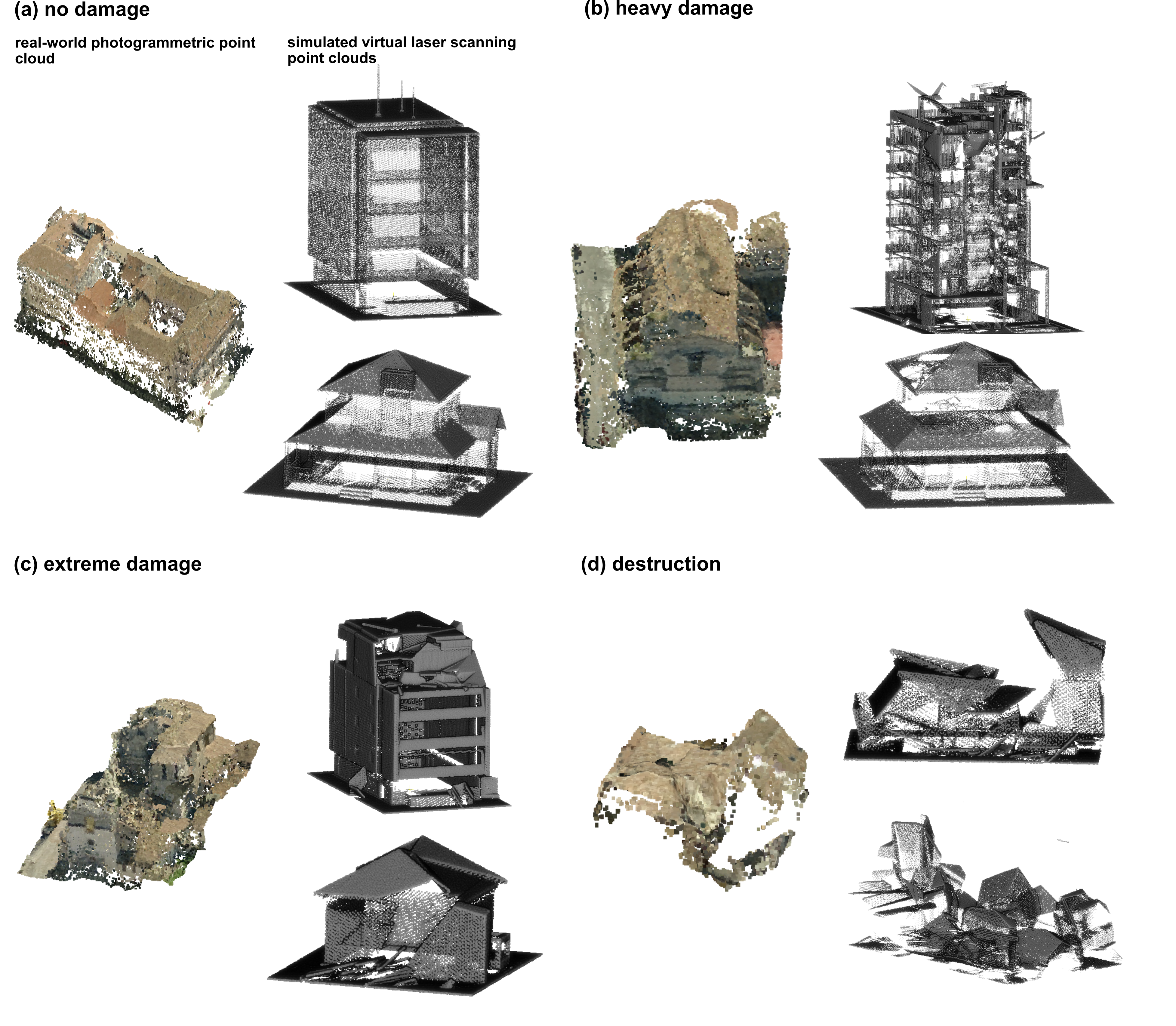}
    \caption{Real-world photogrammetric and simulated virtual laser scanning point clouds for the target damage grades (a) no damage, (b) heavy damage, (c) extreme damage, and (d) destruction.}
    \label{fig:simulated_vs_realworld}
\end{figure}}

\subsection{Object-specific change feature selection}
The difference in relative change of geometric features between real-world ULS and DIM point clouds is visualised in Figure~\ref{fig:change_features_search_rad}. Features with low differences are considered as robust object-specific features which are suitable to classify building damage both in ULS and DIM point clouds (cf. section~\ref{sec:methods-features}). In this study we consider features with less than~10\% difference between ULS and DIM point clouds as this has shown to provide a good compromise between the similarity of geometric change between ULS and DIM point clouds and the number of change features available for damage assessment. The finally selected robust features are: Planarity, surface variation, point density, number of neighbours, surface density, volume density, roughness, z rank, z range and the normal vector. 
The remaining change features (cf.~section~\ref{sec:methods-features}) show strong ($\geq$~15\%) differences between ULS and DIM point clouds and are therefore not considered suitable for our approach. 
\textbf{\begin{figure}[]
    \centering
    \includegraphics[]{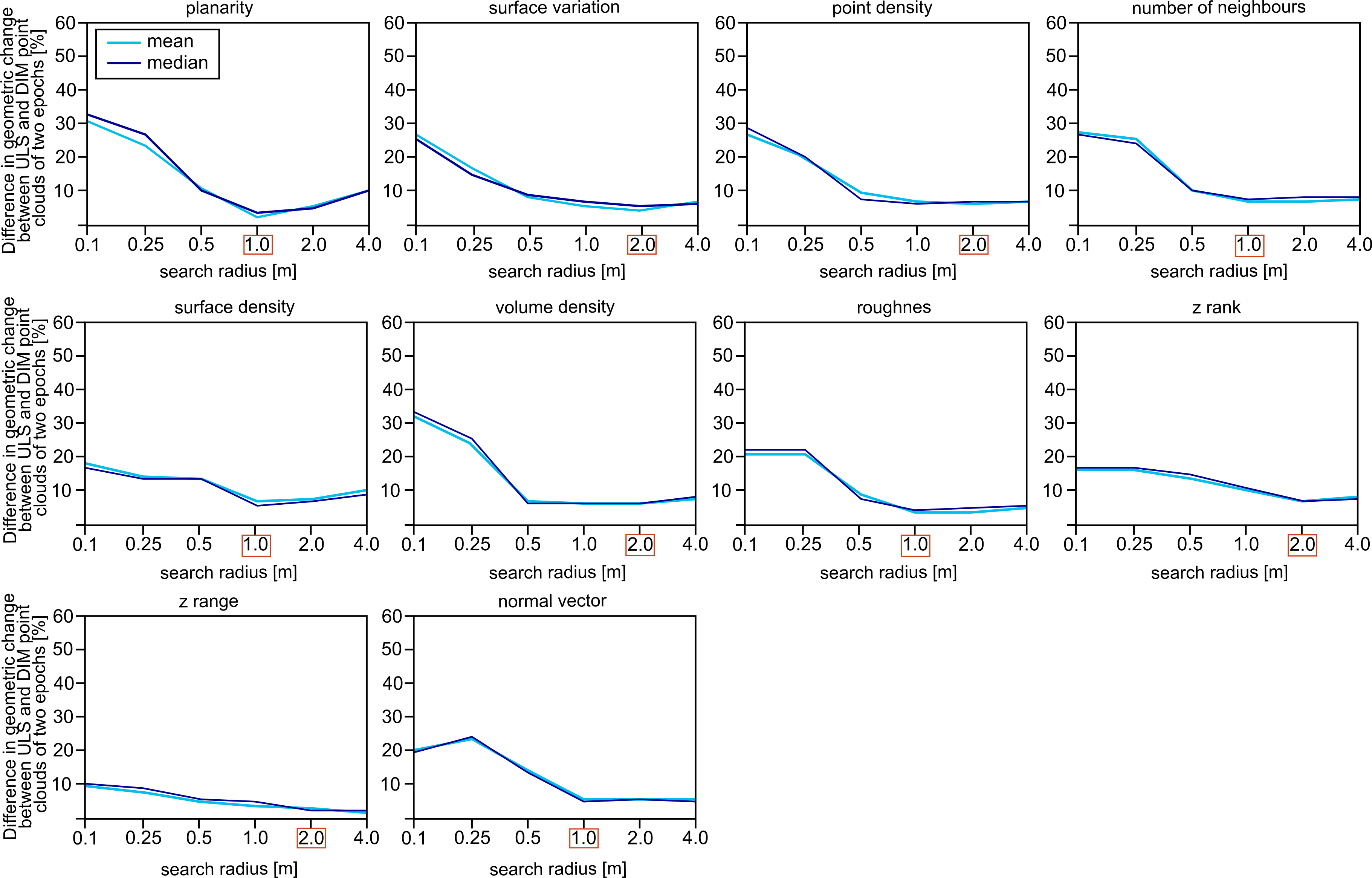}
    \caption{Robust object-specific change features which show less than~10\% difference between ULS and DIM point clouds of two epochs in our experimental investigation. Red boxes indicate the local neighbourhood size for which differences are lowest. These features show to provide a good compromise between the similarity of geometric change between both sources of input point clouds and the number of change features available for damage assessment. They are therefore considered for the random forest-based damage classification.}
    \label{fig:change_features_search_rad}
\end{figure}}
\textbf{\begin{figure}[]
    \centering
    \includegraphics[scale=0.8]{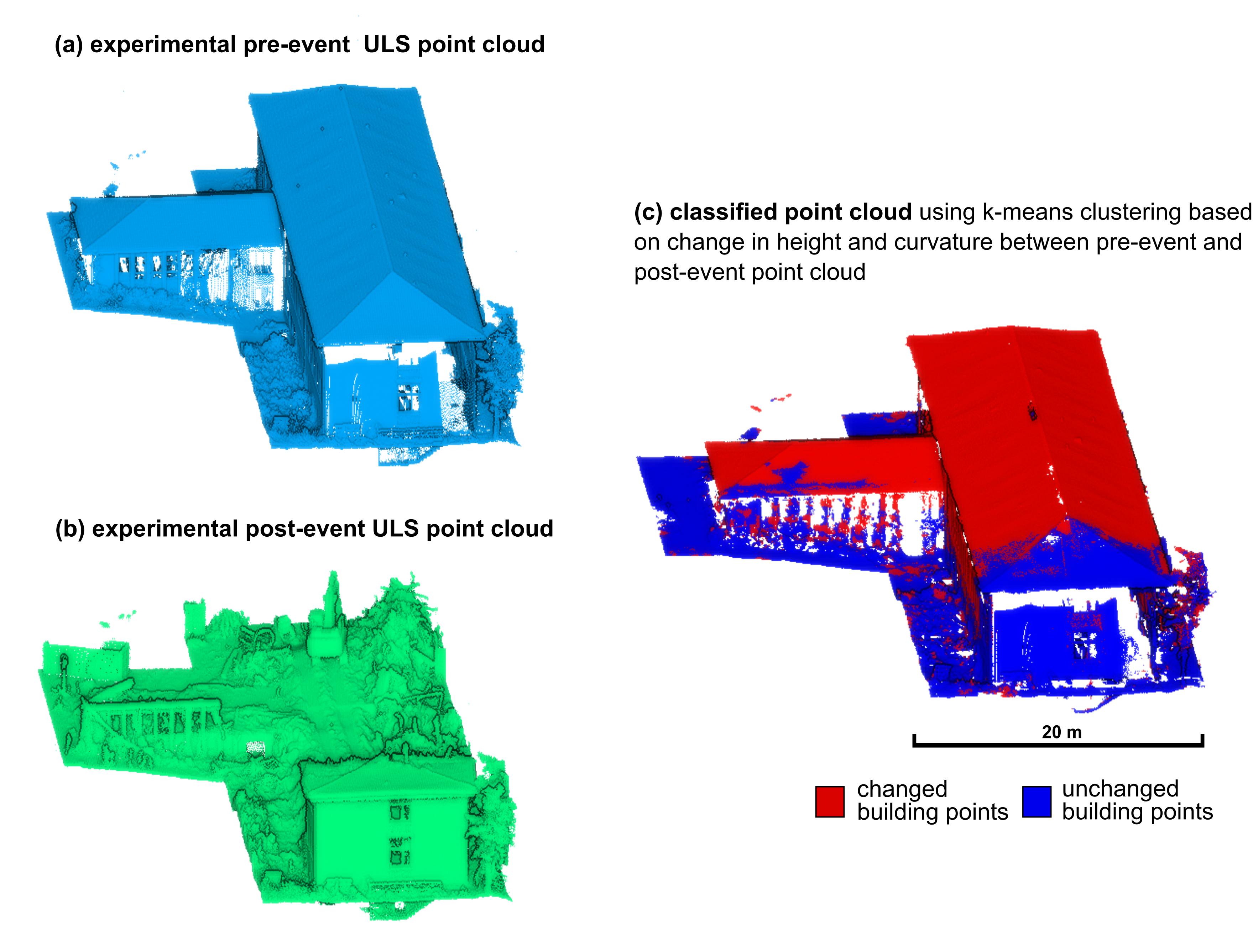}
    \caption{(a)~Pre-event and (b)~post-event UAV-borne laser scanning (ULS) point clouds of the experimental building and (c)~classified changed (red) and unchanged (blue) building parts by k-means clustering. Changed building parts are input for the extraction of object-specific change features as input for the random forest classifier to assess different damage grades.}
    \label{fig:clustering}
\end{figure}}
\subsection{Coarse extraction of changed building parts}
\label{sec:results_clustering}
Results of the k-means clustering to separate changed from unchanged building points are shown by example in Figure~\ref{fig:clustering}. 
This confirms that the clustering based on change in height and curvature performs suitably and therefore is appropriate to filter changed building points for the classification.

\subsection{Evaluation of classifier performances}
\label{sec:results_clf_eval}
The results of all metrics used to quantitatively evaluate the trained models are given in Table~\ref{tab:evaluation_clf} and Figure~\ref{fig:clf_combined}. When applied to the real-world evaluation dataset, the VLS~generic classifier (Figure~\ref{fig:clf_combined}(a)) yields the highest classification accuracy for no damage (overall accuracy:~95.1\%; F1 score~91.7\%), followed by destruction (overall accuracy:~93.6\%; F1 score: 88.9\%),  extreme damage (overall accuracy:~92.0\%; F1 score: 83.3\%), and heavy damage (overall accuracy:~92.8\%; F1 score:~78.1\%).
The VLS region-specific classifier (Figure~\ref{fig:clf_combined}(b)) performs best for the assessment of destruction (overall accuracy:~92.0\%; F1 score~86.5\%), followed by extreme damage (overall accuracy:~92.0\%; F1 score:~83.3\%), heavy damage (overall accuracy:~92.0\%; F1 score:~76.9\%), and no damage (overall accuracy:~92.0\%; F1 score:~76.2\%). 
The VLS~generic + real-world DIM classifier (Figure~\ref{fig:clf_combined}(c)) achieves best performances in the classification of no damage (overall accuracy:~94.4\%; F1 score~90.1\%), followed by destruction (overall accuracy:~92.0\%; F1 score:~86.1\%), extreme damage (overall accuracy:~92.0\%; F1 score:~83.9\%), and heavy damage (overall accuracy:~91.2\%; F1 score:~73.1\%).
The real-world DIM classifier (Figure~\ref{fig:clf_combined} (d)) yields best classification results for no damage (overall accuracy:~96.8\%; F1 score:~94.6\%), followed by destruction (overall accuracy:~93.6\%; F1 score:~89.2\%), heavy damage (overall accuracy:~93.6\%; F1~score: 79.0\%), and extreme damage (overall accuracy:~92.0\%; F1 score:~83.9\%).

\begin{table}[]
\small
\centering
\caption{Accuracy measures for the trained random forest classifiers (VLS~generic, VLS~region-specific, VLS~generic + real-world~DIM, and real-world DIM) for multi-class damage classification of 125~buildings in the real-world~DIM dataset of L'Aquila, Italy. The performance is evaluated for each target damage grade separately as a binary case. Moreover, the overall capacity of the classifiers to correctly separate damaged buildings (any degree of damage) from undamaged buildings is evaluated.}

\begin{tabular}{lccccc}
                                      & \textbf{\begin{tabular}[c]{@{}c@{}}All damage\\ grades\end{tabular}} & \textbf{\begin{tabular}[c]{@{}c@{}}No \\ damage\end{tabular}} & \textbf{\begin{tabular}[c]{@{}c@{}}Heavy \\ damage\end{tabular}} & \textbf{\begin{tabular}[c]{@{}c@{}}Extreme\\ damage\end{tabular}} & \textbf{Destruction} \\ \hline
\textbf{VLS generic}                  &                                                                      &                                                               &                                                                  &                                                                   &                      \\
Overall accuracy                      & \textbf{95.12}                                                       & \textbf{95.12}                                                & \textbf{92.80}                                                   & \textbf{92.00}                                                    & \textbf{93.60}       \\
Precision                             & 1.0                                                                  & 84.62                                                         & 72.73                                                            & 78.13                                                             & 91.43                \\
Recall                                & 84.62                                                                & 1.0                                                           & 84.21                                                            & 89.29                                                             & 86.49                \\
F1 score                              & \textbf{91.67}                                                       & \textbf{91.67}                                                & \textbf{78.05}                                                   & \textbf{83.33}                                                    & \textbf{88.89}       \\ \hline
\textbf{VLS region-specific}          &                                                                      &                                                               &                                                                  &                                                                   &                      \\
Overall accuracy                      & 94.40                                                                & 92.00                                                         & 92.00                                                            & 92.00                                                             & 92.00                \\
Precision                             & 82.05                                                                & 84.21                                                         & 69.57                                                            & 78.13                                                             & 82.05                \\
Recall                                & 1.0                                                                  & 69.57                                                         & 84.21                                                            & 89.29                                                             & 91.43                \\
F1 score                              & 90.14                                                                & 76.19                                                         & 76.91                                                            & 83.33                                                             & 86.49                \\ \hline
\textbf{VLS generic + real-world DIM} &                                                                      &                                                               &                                                                  &                                                                   &                      \\
Overall accuracy                      & 94.40                                                                & 94.40                                                         & 91.20                                                            & 92.00                                                             & 92.00                \\
Precision                             & 82.05                                                                & 1.0                                                           & 68.18                                                            & 81.25                                                             & 83.78                \\
Recall                                & 1.0                                                                  & 82.05                                                         & 78.95                                                            & 86.67                                                             & 88.57                \\
F1 score                              & 90.14                                                                & 90.14                                                         & 73.17                                                            & 83.87                                                             & 86.11                \\ \hline
\textbf{Real-world DIM}               &                                                                      &                                                               &                                                                  &                                                                   &                      \\
Overall accuracy                      & \textbf{96.80}                                                       & \textbf{96.80}                                                & \textbf{93.60}                                                   & \textbf{92.00}                                                    & \textbf{93.60}       \\
Precision                             & 1.0                                                                  & 89.74                                                         & 78.95                                                            & 81.25                                                             & 84.62                \\
Recall                                & 89.74                                                                & 1.0                                                           & 78.95                                                            & 86.67                                                             & 94.29                \\
F1 score                              & \textbf{94.59}                                                       & \textbf{94.59}                                                & \textbf{78.95}                                                   & \textbf{83.87}                                                    & \textbf{89.19}      
\end{tabular}
\label{tab:evaluation_clf}
\end{table}

Inter-class confusions for all classifiers mainly occur between neighbouring damage grades. For example, buildings with extreme damage are misclassified as heavy damage or destruction. Only buildings with no damage are partly misclassified as heavy damage or vice versa by all classifiers, which are significantly different degrees of damage. For one of these miss-classified buildings, a visual inspection of the point clouds reveals occlusion effects in the area of damage, which results from the more narrow built structure in the region-specific simulated scene. For the other buildings, however, close inspection does not show occlusion effects but the class probabilities of no damage and heavy damage in these cases are similarly high which suggests, that damage patterns of these building objects are not significantly different from no damage.

Using generic simulated training data yields good classification results for all target classes with overall classification accuracies between 92.0\% and 95.1\%, and F1 scores ranging between 78.1\% and 91.7\%. Using region-specific simulated instead of generic simulated training data does not strongly reduce inter-class confusion or increase the completeness of detected damaged buildings (+3\%), neither does the use of real-world region-specific training data (+6\% increase in completeness of detected damaged buildings). This implies that our model trained purely on generic simulated data has a high transferability to unseen regions and that the benefit of adding site-specific real-world training data is low for the classification task at hand.

The performance of the VLS generic classifier to detect damage, i.e. binary classification to separate damaged from undamaged buildings, is strong with an overall accuracy of~84.6\%. Using real-world region-specific training data achieves only slightly higher overall accuracies~(89.7\%).

Simulated or real-world region-specific training data does hence not considerably increase classification accuracies neither with respect to the classification of multiple damage grades, nor with respect to the detection of damaged buildings. We attribute this to the fact that for the damage grades considered in our study, damage patterns do not vary considerably for different building types and built structure. Change features learned from the generic simulated training dataset hence generalise appropriately to be used for damage assessment in datasets with different site characteristics. These results support our hypothesis that the transfer of a supervised machine learning model trained purely on simulated, non-region specific training data to an unseen real-world dataset with different site characteristics is a valid approach for the use case of timely damage assessment for earthquake response. Although the degree of damage is not increasing linearly but rather exponentially with increasing damage grades, inter-class confusion for all classifiers mostly occurs between neighbouring damage grades and not between damage grades far from each other. We therefore assume that the damage catalogue as a descriptive framework is not exclusively relating to geometric change in the point cloud. Consequently, the model might learn certain geometric representations of damage grades differently from how expert analysts would categorise them. Moreover, except from one building, all damaged buildings are correctly detected as damaged. This is an important result, as applications can rely to identify damaged buildings with high completeness. The exact degree of damage may not be as essential in this first step but it may already be useful to derive a tendency towards rather high or low degree of damage. 

\textbf{\begin{figure}[]
    \centering
    \includegraphics[scale=0.9]{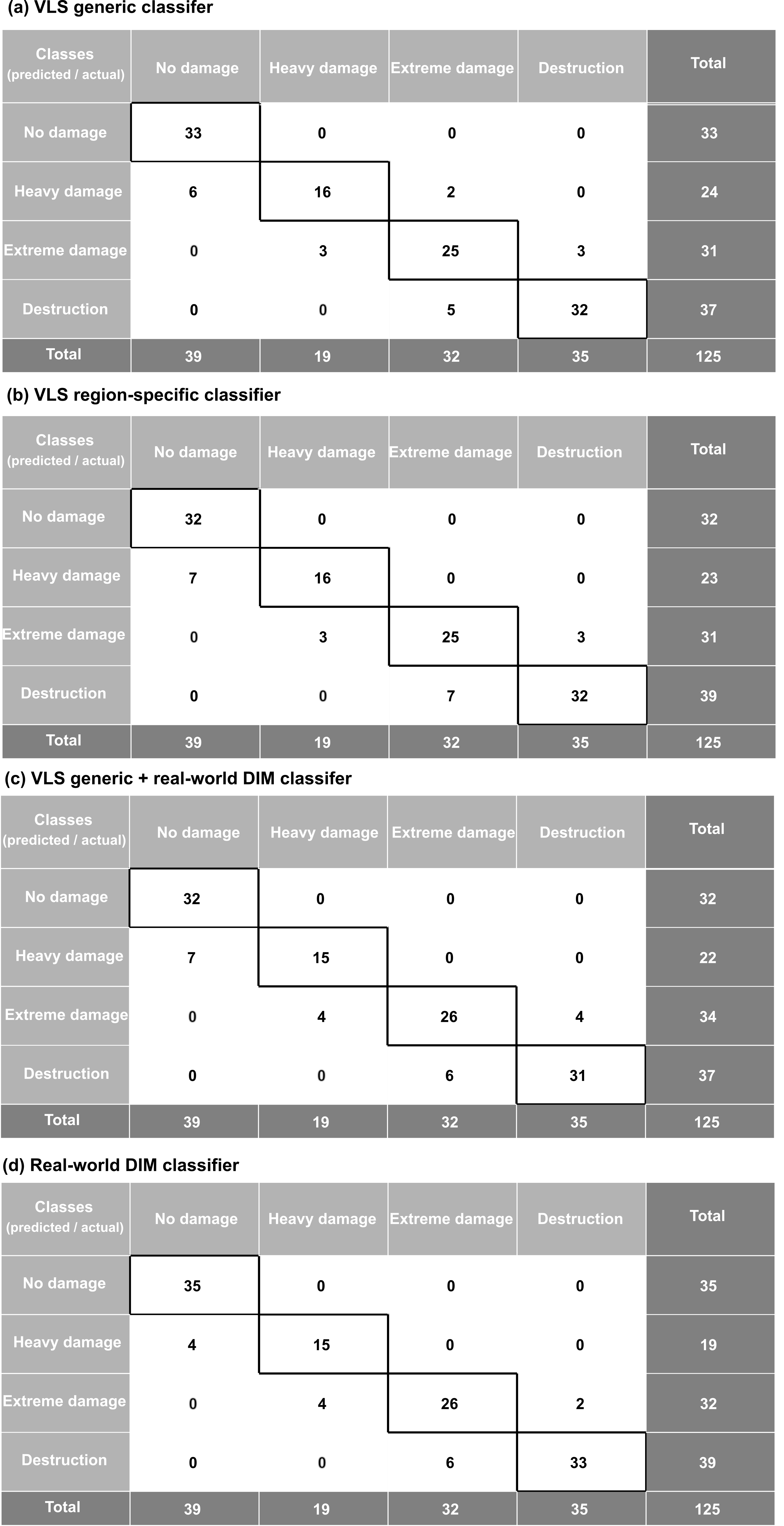}
    \caption{Confusion matrices for the trained random forest classifiers (a) VLS generic, (b) VLS region-specific, (c) VLS generic + real-world DIM, and (d) real-world DIM for multi-class damage classification of 125 buildings in the real-world photogrammetric dataset of L'Aquila, Italy.}
    \label{fig:clf_combined}
\end{figure}}

\section{Discussion}
\label{sec:discussion}
\subsection{Transferability of the method with respect to the source of input point clouds}
Our approach achieves transferability of input point clouds from different sources (laser scanning and photogrammetry) for training and application of the random forest classifier. More specifically, we are able to transfer the classifier trained on simulated UAV-borne laser scanning point clouds to classify damage in real-world UAV-borne DIM point clouds. We achieve this through the identification of object-specific geometric change features, which show to be representative for the investigated damage grades both when using ULS and DIM point clouds. Handling multi-source point clouds is especially relevant when training models with simulated data, due to the lack of available tools to simulate photogrammetric point clouds. Therefore, training with simulated data has to be based on simulated laser scanning point clouds, whereas the application of the model might be based on DIM point clouds.
Our approach could be extended to consider multi-modal data also with respect to the pre-event and post-event input point clouds used to derive change between two epochs, such as performed in \cite{dai_2020}. In practice, point cloud data of built areas is often available from different sensor systems and acquisition strategies, such as ours, or may be generated from existing 3D city models. Consequently, the confident assessment of structural building damage requires  methods and strategies which can extract representative change features from heterogeneous pre-event and post-event point clouds \citep{zhang_et_al_2021}.

\subsection{Geographic transferability of the method}
A further important strength of our approach is the geographic transferability of a trained classifier to datasets from unseen regions. We found that the random forest classifier trained on generic (i.e., not region-specific) simulated building point clouds achieves high classification accuracies in the real-world dataset, which are comparable to those achieved in studies training on region-specific inputs, such as \cite{deGelis_2021}. Using simulated region-specific building point clouds with damage characteristics more similar to the real-world building point clouds does not strongly improve the classification results (increase of overall accuracy $<$~2\%; increase of F1 score: $<$~2\%), neither does training purely based on real-world region-specific data (increase of overall accuracy $<$~2\%; increase of F1 score: $<$~3\%).

Model transferability is still a major challenge for the assessment of binary or multi-class building damage. For example, \cite{vetrivel_2018} found that the geographic transferability of their supervised model based on a CNN and 3D point cloud features is limited already when scene characteristics vary slightly. 
In our study, we achieve model transferability through the integration of domain knowledge in the process of training data generation. Using the concept of a damage catalogue developed in \cite{Kohns_et_al_2021a}, we are able to identify damage patterns which characterise the target damage grades across different geographic regions, and model them into the virtual 3D building models. Therefore, our model is trained on geometric change features which generalise adequately to discriminate target damage grades in point clouds of different geographic regions. 

This contribution of our method is of utmost relevance for practical use. In contrast to a computational estimation of building damage based on pre-calculated fragility functions \citep{kohns_2022b} it allows assessing damage directly for the specific buildings in the affected area, and to directly applying the pre-trained model for damage classification in a real earthquake event as soon as post-event UAV data is available.

\subsection{Automatic generation of large 3D training datasets}
Simulated training data in this study comprises the manual modelling of various damage patterns into 3D building models, and the annotation of buildings with damage labels. As such training data can be generated to train machine learning models before an actual earthquake event occurs, the trained models can be directly applied to classify damage in real-world datasets acquired after an earthquake occurs. This saves valuable time for rescue and remediation actions to set in \citep{Kohns_et_al_2021a}. Future approaches of UAV damage assessment in earthquake response might integrate databases of existing damaged and undamaged 3D building models. Such databases could integrate both building models generated from real-world city models \citep[e.g.,][]{Zhihang_2018} and synthetically generated building models from other sources, including virtual laser scanning.
In the field of forestry, for example, the Python software package pytreedb \citep{pytreedb} provides an open object-based library for laser scanning point clouds of trees with a simple interface to store and share the tree objects. A similar approach could be used to store and query labelled 3D building models in an open database. The database could even be automatically connected with the laser scanning simulation module to assemble a multitude of different scenes where building objects can be flexibly interchanged. Simulated building point clouds might then be stored with the respective building object in the database. Such an approach considerably increases the degree of automation in the process of training data generation. It also provides great flexibility in assembling input scenes for laser scanning simulation. The approach can thereby be especially valuable for classification methods with high demands for training data, such as Deep Learning \citep{deGelis_2021}.
While in our approach we classify damage on building level, it generally offers flexibility with respect to the spatial unit of  extraction of changed building part and subsequent damage classification. Instead of a whole building, the approach might in the future also be tested to assess damage only for partial buildings which represent a coherent unit that is of interest for damage classification for a specific use case. 

\subsection{Automatic assessment of lower damage grades in UAV-borne point clouds}
In our method, we use a geometric approach for the assessment of structural building damage through change in pre- and post-event 3D~UAV-borne point clouds. We therein focus on the assessment of higher damage grades (heavy damage, extreme damage, destruction). These are usually geometrically resolved in UAV-borne point clouds acquired over large areas during an earthquake event. When interpreting classification results, it should be considered that buildings classified as non-damaged by our approach might be slightly or moderately damaged, but cannot be recognised as such from the point cloud data due to more subtle damage patterns. 
While high damage grades are most relevant to support the coordination of immediate rescue actions on site in the very first hours after an earthquake event, a timely assessment of lower damage grades is also of great relevance, especially because these buildings are prone to severe damage in aftershock events. 

Other UAV-borne acquisition strategies are needed to meet the requirements for the assessment of lower damage grades. Most importantly, the spatial resolution of derived point clouds would need to geometrically represent typical lower-grade damage patterns, such as cracks of millimeter width. This can be achieved by lower flight altitudes and flight speeds to focus on selected individual buildings at higher detail. Such acquisitions strategies favour high spatial resolution over large spatial coverage of the affected area \citep{meyer_2022}.

To assess both high and low damage grades in a reasonable time frame, a two-step approach could be deployed \citep{kohns_et_al_egu_2021}: Immediately after an earthquake, fleets of UAVs are sent out to capture the entire affected area in overview flights within a few hours. Areas of higher relevance for immediate damage assessment can be prioritised in the acquisition through the integration of an initial damage assessment in the planning of the UAV missions. This initial damage forecast is based on ground motion fields and exposure data including fragility and vulnerability functions, and gives a first estimate on the expected building-specific damage. Point clouds resulting from such overview flights are characterised by medium to low resolution ($>$~0.1~m), and can be input for the automatic and fast assessment of higher damage grades (destruction, extreme damage, heavy damage), as in our method.  Beyond immediate response actions, this first assessment reduces the number of buildings that require a more detailed UAV-borne acquisition to identify potential lower damage grades (slight damage, moderate damage). Hence, only buildings classified as not damaged in the coarse assessment will be considered in detailed UAV acquisitions and the subsequent assessment of damage, which considerably reduces the amount of data to be analysed. Such approaches support local rescue teams during various phases after an earthquake.

\section{Conclusion}
We present a novel approach to automatically classify multi-class structural building damage from multi-temporal point clouds, evaluated on pre- and post-event data of an earthquake event. We evaluate a supervised machine learning model trained on simulated point clouds from virtual UAV-borne laser scanning with respect to its capacity to classify damage grades no damage, heavy damage, extreme damage, and destruction in a real-world photogrammetric dataset. Damage is thereby assessed through an approach that considers change of geometric features between a pre-event and post-event point cloud for the classification of the grades of damage.

Our results reveal transferability with respect to multi-source point clouds used for training (simulated UAV-borne laser scanning) and application (real-world UAV-borne photogrammetry) of the model. This is possible by using a set of robust object-specific change features, which characterise damage patterns both in the simulated and real-world point clouds. 
Consequently, simulated point clouds from virtual laser scanning provide a valuable source of realistic labelled training data for the classification task at hand.

We further achieve geographic transferability of the model trained on simulated point clouds by integrating domain knowledge from earthquake engineering in the generation of realistic simulated training data. This allows training the model on geometric change which characterises the target damage grades across different geographic regions. The assessment of multiple damage grades in the real-world dataset yields high accuracies (overall accuracy:~92.0\%~-~95.1\% ; F1 score:~78.05~-~91.67\%). Classification accuracies only slightly improve when using real-world region-specific training data (overall accuracy:~$<$~2\%, F1 score:~$<$~3\%). The same applies for the binary case of detecting damaged buildings for which the classifier trained on generic simulated training data detects~89.6\% of damaged buildings. Using real-world region-specific training data only slightly increases the detection rate by~+3.1\%.

Given the two aspects of transferability, we conclude that our approach provides a powerful assessment of multi-class structural building damage. We consider it especially relevant for applications where timely information on the damage situation is required, often linked to the situation that sufficient real-world training data is not available.

\newpage

\beginsupplement

\section*{Data statement}
The 3D building models, simulated point clouds and Python scripts used in this article will be openly provided after peer-reviewed publication.

\section*{Acknowledgement}
This work was supported by the Bundesministerium für Bildung und Forschung (BMBF), Federal Ministry of Education and Research under Grant 03G0890 in the frame of the project LOKI and by the Deutsche Forschungsgemeinschaft (DFG), Germany Research Foundation, in the frame of the project VirtuaLearn3D under Grant 496418931. We would like to thank CGR S.p.A. (Compagnia Generale Ripreseaeree) for providing the real-world image data of the city of L'Aquila for deriving the photogrammetric point clouds used in this study.





\bibliographystyle{elsarticle-harv}

\bibliography{main.bib}

\begin{thebibliography}{45}
\expandafter\ifx\csname natexlab\endcsname\relax\def\natexlab#1{#1}\fi
\providecommand{\url}[1]{\texttt{#1}}
\providecommand{\href}[2]{#2}
\providecommand{\path}[1]{#1}
\providecommand{\DOIprefix}{doi:}
\providecommand{\ArXivprefix}{arXiv:}
\providecommand{\URLprefix}{URL: }
\providecommand{\Pubmedprefix}{pmid:}
\providecommand{\doi}[1]{\href{http://dx.doi.org/#1}{\path{#1}}}
\providecommand{\Pubmed}[1]{\href{pmid:#1}{\path{#1}}}
\providecommand{\bibinfo}[2]{#2}
\ifx\xfnm\relax \def\xfnm[#1]{\unskip,\space#1}\fi
\bibitem[{Alzubaidi et~al.(2021)Alzubaidi, Zhang, Humaidi, Al-Dujaili, Duan,
  Al-Shamma, Santamaría, Fadhel, Al-Amidie and Farhan}]{alzubaidi_2021}
\bibinfo{author}{Alzubaidi, L.}, \bibinfo{author}{Zhang, J.},
  \bibinfo{author}{Humaidi, A.J.}, \bibinfo{author}{Al-Dujaili, A.},
  \bibinfo{author}{Duan, Y.}, \bibinfo{author}{Al-Shamma, O.},
  \bibinfo{author}{Santamaría, J.}, \bibinfo{author}{Fadhel, M.A.},
  \bibinfo{author}{Al-Amidie, M.}, \bibinfo{author}{Farhan, L.},
  \bibinfo{year}{2021}.
\newblock \bibinfo{title}{Review of deep learning: concepts, {CNN}
  architectures, challenges, applications, future directions}.
\newblock \bibinfo{journal}{Journal of Big Data} \bibinfo{volume}{8},
  \bibinfo{pages}{1--74}.
\newblock \DOIprefix\doi{10.1186/s40537-021-00444-8}.
\bibitem[{Awrangjeb et~al.(2015)Awrangjeb, Fraser and Lu}]{awrangjeb_2015}
\bibinfo{author}{Awrangjeb, M.}, \bibinfo{author}{Fraser, C.S.},
  \bibinfo{author}{Lu, G.}, \bibinfo{year}{2015}.
\newblock \bibinfo{title}{Building change detection from {LiDAR} point cloud
  data based on connected component analysis}.
\newblock \bibinfo{journal}{ISPRS Annals of the Photogrammetry, Remote Sensing
  and Spatial Information Sciences} \bibinfo{volume}{II-3/W5},
  \bibinfo{pages}{1--8}.
\newblock \DOIprefix\doi{10.5194/isprsannals-II-3-W5-393-2015}.
\bibitem[{Besl and McKay(1992)}]{besl_mckay_1992}
\bibinfo{author}{Besl, P.}, \bibinfo{author}{McKay, N.D.},
  \bibinfo{year}{1992}.
\newblock \bibinfo{title}{A method for registration of 3-{D} shapes}.
\newblock \bibinfo{journal}{IEEE Transactions on Pattern Analysis and Machine
  Intelligence} \bibinfo{volume}{14}, \bibinfo{pages}{239--256}.
\newblock \DOIprefix\doi{10.1109/34.121791}.
\bibitem[{{Blender Online Community}(2018)}]{blender_2018}
\bibinfo{author}{{Blender Online Community}}, \bibinfo{year}{2018}.
\newblock \bibinfo{title}{{B}lender - {A} {3D} modelling and rendering
  package.} \URLprefix \url{http://www.blender.org}.
\bibitem[{Breiman(2001)}]{breiman_random_2001}
\bibinfo{author}{Breiman, L.}, \bibinfo{year}{2001}.
\newblock \bibinfo{title}{Random {Forests}}.
\newblock \bibinfo{journal}{Machine Learning} \bibinfo{volume}{45 (1)},
  \bibinfo{pages}{5--32}.
\newblock \DOIprefix\doi{10.1023/A:1010933404324}.
\bibitem[{Dai et~al.(2020)Dai, Zhang and Lin}]{dai_2020}
\bibinfo{author}{Dai, C.}, \bibinfo{author}{Zhang, Z.}, \bibinfo{author}{Lin,
  D.}, \bibinfo{year}{2020}.
\newblock \bibinfo{title}{An object-based bidirectional method for integrated
  building extraction and change detection between multimodal point clouds}.
\newblock \bibinfo{journal}{Remote Sensing} \bibinfo{volume}{12},
  \bibinfo{pages}{1--23}.
\newblock \DOIprefix\doi{10.5194/10.3390/rs12101680}.
\bibitem[{Duarte et~al.(2020)Duarte, Nex, Kerle and Vosselman}]{duarte_2020}
\bibinfo{author}{Duarte, D.}, \bibinfo{author}{Nex, F.},
  \bibinfo{author}{Kerle, N.}, \bibinfo{author}{Vosselman, G.},
  \bibinfo{year}{2020}.
\newblock \bibinfo{title}{Detection of seismic façade damages with
  multi-temporal oblique aerial imagery}.
\newblock \bibinfo{journal}{GIScience \& Remote Sensing} \bibinfo{volume}{57},
  \bibinfo{pages}{670--686}.
\newblock \DOIprefix\doi{10.1080/15481603.2020.1768768}.
\bibitem[{{Earthquake Engineering Research Institute}(2009)}]{eer_2009}
\bibinfo{author}{{Earthquake Engineering Research Institute}},
  \bibinfo{year}{2009}.
\newblock \bibinfo{title}{Learning from earthquakes: The mw 6.3 {A}bruzzo,
  {I}taly, earthquake of {A}pril 6, 2009.}
\newblock \bibinfo{journal}{EERI Special Earthquake Report, June, EERI,
  Oakland, CA} , \bibinfo{pages}{1--12.{ }}\URLprefix
  \url{https://www.reluis.it/doc/pdf/Aquila/EERI_L_Aquila_report.pdf}.
\bibitem[{Fernandez~Galarreta et~al.(2015)Fernandez~Galarreta, Kerle and
  Gerke}]{galareta_2015}
\bibinfo{author}{Fernandez~Galarreta, J.}, \bibinfo{author}{Kerle, N.},
  \bibinfo{author}{Gerke, M.}, \bibinfo{year}{2015}.
\newblock \bibinfo{title}{{UAV}-based urban structural damage assessment using
  object-based image analysis and semantic reasoning}.
\newblock \bibinfo{journal}{Natural Hazards and Earth System Sciences}
  \bibinfo{volume}{15}, \bibinfo{pages}{1087--1101}.
\newblock \DOIprefix\doi{10.5194/nhess-15-1087-2015}.
\bibitem[{Gastellu-Etchegorry et~al.(2016)Gastellu-Etchegorry, Yin, Lauret,
  Grau, Rubio, Cook, Morton and Sun}]{gastellu_2021}
\bibinfo{author}{Gastellu-Etchegorry, J.P.}, \bibinfo{author}{Yin, T.},
  \bibinfo{author}{Lauret, N.}, \bibinfo{author}{Grau, E.},
  \bibinfo{author}{Rubio, J.}, \bibinfo{author}{Cook, B.D.},
  \bibinfo{author}{Morton, D.C.}, \bibinfo{author}{Sun, G.},
  \bibinfo{year}{2016}.
\newblock \bibinfo{title}{Simulation of satellite, airborne and terrestrial
  {LiDAR} with {DART} ({I}): Waveform simulation with quasi-{M}onte {C}arlo ray
  tracing}.
\newblock \bibinfo{journal}{Remote Sensing of Environment}
  \bibinfo{volume}{184}, \bibinfo{pages}{418--435}.
\newblock \DOIprefix\doi{10.1016/j.rse.2016.07.010}.
\bibitem[{{Geo-Engineering Extreme Events Reconnaissance}(2009)}]{geer_2009}
\bibinfo{author}{{Geo-Engineering Extreme Events Reconnaissance}},
  \bibinfo{year}{2009}.
\newblock \bibinfo{title}{Preliminary report on the seismological and
  geotechnical aspects of the april 6 2009 l’aquila earthquake in central
  italy (version 2.0)}.
\newblock \bibinfo{journal}{GEER Association Report No. GEER-016, September
  2009} , \bibinfo{pages}{1--166.{ }}\URLprefix
  \url{https://www.academia.edu/67093517/Preliminary_Report_on_the_Seismological_and_Geotechnical_Aspects_of_the_April_6_2009_LAquila_Earthquake_in_Central_Italy_Version_2_0_GEER_Association_Report}.
\bibitem[{de~Gélis et~al.(2021)de~Gélis, Lefèvre and
  Corpetti}]{deGelis_2021}
\bibinfo{author}{de~Gélis, I.}, \bibinfo{author}{Lefèvre, S.},
  \bibinfo{author}{Corpetti, T.}, \bibinfo{year}{2021}.
\newblock \bibinfo{title}{Change detection in urban point clouds: An
  experimental comparison with simulated {3D} datasets}.
\newblock \bibinfo{journal}{Remote Sensing} \bibinfo{volume}{13},
  \bibinfo{pages}{1--29}.
\newblock \DOIprefix\doi{10.3390/rs13132629}.
\bibitem[{Hildebrand et~al.(2022)Hildebrand, Schulz, Richter and
  D\"ollner}]{hildebrand_2022}
\bibinfo{author}{Hildebrand, J.}, \bibinfo{author}{Schulz, S.},
  \bibinfo{author}{Richter, R.}, \bibinfo{author}{D\"ollner, J.},
  \bibinfo{year}{2022}.
\newblock \bibinfo{title}{Simulating {LiDAR} to create training data for
  machine learning on 3d point clouds.}
\newblock \bibinfo{journal}{ISPRS Annals of the Photogrammetry, Remote Sensing
  and Spatial Information Sciences} \bibinfo{volume}{X-4/W2-2022},
  \bibinfo{pages}{105--112.{ }}.
\newblock \DOIprefix\doi{10.5194/isprs-annals-X-4-W2-2022-105-2022}.
\bibitem[{Huang et~al.(2019)Huang, Sun, Zhang, Hao, Zhang, Ren and
  Ma}]{huang_2019}
\bibinfo{author}{Huang, H.}, \bibinfo{author}{Sun, G.}, \bibinfo{author}{Zhang,
  X.}, \bibinfo{author}{Hao, Y.}, \bibinfo{author}{Zhang, A.},
  \bibinfo{author}{Ren, J.}, \bibinfo{author}{Ma, H.}, \bibinfo{year}{2019}.
\newblock \bibinfo{title}{Combined multiscale segmentation convolutional neural
  network for rapid damage mapping from postearthquake very high-resolution
  images}.
\newblock \bibinfo{journal}{Journal of Applied Remote Sensing}
  \bibinfo{volume}{13}, \bibinfo{pages}{1--15}.
\newblock \DOIprefix\doi{10.1117/1.JRS.13.022007}.
\bibitem[{Höfle et~al.(2009)Höfle, Mücke, Dutter, Rutzinger and
  Dorninger}]{hoefle_2009}
\bibinfo{author}{Höfle, B.}, \bibinfo{author}{Mücke, W.},
  \bibinfo{author}{Dutter, M.}, \bibinfo{author}{Rutzinger, M.},
  \bibinfo{author}{Dorninger, P.}, \bibinfo{year}{2009}.
\newblock \bibinfo{title}{Detection of building regions using airborne lidar: a
  new combination of raster and point cloud based gis methods}.
\newblock \bibinfo{journal}{Proceedings of the {T}hird {G}eoinformatics {F}orum
  Salzburg: Geoinformatics on stage, July 7-10, 2009.} ,
  \bibinfo{pages}{66--75}.
\bibitem[{Höfle et~al.(2022)Höfle, Qu, Winiwarter, Weiser, Zahs, Schäfer and
  Fassnacht}]{pytreedb}
\bibinfo{author}{Höfle, B.}, \bibinfo{author}{Qu, J.},
  \bibinfo{author}{Winiwarter, L.}, \bibinfo{author}{Weiser, H.},
  \bibinfo{author}{Zahs, V.}, \bibinfo{author}{Schäfer, J.},
  \bibinfo{author}{Fassnacht, F.E.}, \bibinfo{year}{2022}.
\newblock \bibinfo{title}{pytreedb: library for point clouds of tree vegetation
  objects} \URLprefix \url{https://github.com/3dgeo-heidelberg/pytreedb}.
\bibitem[{Kalantar et~al.(2020)Kalantar, Ueda, Al-Najjar and
  Halin}]{kalantar_2020}
\bibinfo{author}{Kalantar, B.}, \bibinfo{author}{Ueda, N.},
  \bibinfo{author}{Al-Najjar, H.A.H.}, \bibinfo{author}{Halin, A.A.},
  \bibinfo{year}{2020}.
\newblock \bibinfo{title}{Assessment of convolutional neural network
  architectures for earthquake-induced building damage detection based on pre-
  and post-event orthophoto images}.
\newblock \bibinfo{journal}{Remote Sensing} \bibinfo{volume}{12},
  \bibinfo{pages}{1--22}.
\newblock \DOIprefix\doi{10.3390/rs12213529}.
\bibitem[{Kerle et~al.(2020)Kerle, Nex, Gerke, Duarte and
  Vetrivel}]{kerle_2020}
\bibinfo{author}{Kerle, N.}, \bibinfo{author}{Nex, F.}, \bibinfo{author}{Gerke,
  M.}, \bibinfo{author}{Duarte, D.}, \bibinfo{author}{Vetrivel, A.},
  \bibinfo{year}{2020}.
\newblock \bibinfo{title}{{UAV}-based structural damage mapping: A review}.
\newblock \bibinfo{journal}{ISPRS International Journal of Geo-Information}
  \bibinfo{volume}{9}, \bibinfo{pages}{1--23}.
\newblock \DOIprefix\doi{10.3390/ijgi9010014}.
\bibitem[{Kharroubi et~al.(2022)Kharroubi, Poux, Ballouch, Hajji and
  Billen}]{kharroubi_2022}
\bibinfo{author}{Kharroubi, A.}, \bibinfo{author}{Poux, F.},
  \bibinfo{author}{Ballouch, Z.}, \bibinfo{author}{Hajji, R.},
  \bibinfo{author}{Billen, R.}, \bibinfo{year}{2022}.
\newblock \bibinfo{title}{Three dimensional change detection using point
  clouds: A review}.
\newblock \bibinfo{journal}{Geomatics} \bibinfo{volume}{2},
  \bibinfo{pages}{457--485}.
\newblock \DOIprefix\doi{10.3390/geomatics2040025}.
\bibitem[{Kohns and Stempniewski(2021)}]{Kohns_et_al_2021a}
\bibinfo{author}{Kohns, J.}, \bibinfo{author}{Stempniewski, L.},
  \bibinfo{year}{2021}.
\newblock \bibinfo{title}{Classification of earthquake-induced building damage
  using innovative methods}.
\newblock \bibinfo{journal}{IABSE Congress: Structural Engineering for Future
  Societal Needs} , \bibinfo{pages}{1366--1374.{
  }}\DOIprefix\doi{10.2749/ghent.2021.1366}.
\bibitem[{Kohns et~al.(2022a)Kohns, Stempniewski and Stark}]{kohns_2022}
\bibinfo{author}{Kohns, J.}, \bibinfo{author}{Stempniewski, L.},
  \bibinfo{author}{Stark, A.}, \bibinfo{year}{2022}a.
\newblock \bibinfo{title}{Development of damage catalogues for visual
  assessment of buildings in the event of an earthquake}.
\newblock \bibinfo{journal}{Bauingenieur} \bibinfo{volume}{97},
  \bibinfo{pages}{403--–412}.
\newblock \DOIprefix\doi{10.37544/0005-6650-2022-12-39}.
\bibitem[{Kohns et~al.(2022b)Kohns, Stempniewski and Stark}]{kohns_2022b}
\bibinfo{author}{Kohns, J.}, \bibinfo{author}{Stempniewski, L.},
  \bibinfo{author}{Stark, A.}, \bibinfo{year}{2022}b.
\newblock \bibinfo{title}{Fragility functions for reinforced concrete
  structures based on multiscale approach for earthquake damage criteria}.
\newblock \bibinfo{journal}{Buildings} \bibinfo{volume}{12},
  \bibinfo{pages}{1--17}.
\newblock \DOIprefix\doi{10.3390/buildings12081253}.
\bibitem[{Kohns et~al.(2021)Kohns, Zahs, Ullah, Schorlemmer, Nievas, Glock,
  Meyer, Stempniewski, Zipf and H\"ofle}]{kohns_et_al_egu_2021}
\bibinfo{author}{Kohns, J.}, \bibinfo{author}{Zahs, V.},
  \bibinfo{author}{Ullah, T.}, \bibinfo{author}{Schorlemmer, D.},
  \bibinfo{author}{Nievas, C.}, \bibinfo{author}{Glock, K.},
  \bibinfo{author}{Meyer, F.}, \bibinfo{author}{Stempniewski, L.and~Herfort,
  B.}, \bibinfo{author}{Zipf, A.}, \bibinfo{author}{H\"ofle, B.},
  \bibinfo{year}{2021}.
\newblock \bibinfo{title}{Innovative methods for earthquake damage detection
  and classification using airborne observation of critical infrastructures
  (project {LOKI})}.
\newblock \bibinfo{journal}{EGU General Assembly 2021} , \bibinfo{pages}{1--1.{
  }}\DOIprefix\doi{10.5194/egusphere-egu21-2712}.
\bibitem[{Lague et~al.(2013)Lague, Brodu and Leroux}]{lague_accurate_2013}
\bibinfo{author}{Lague, D.}, \bibinfo{author}{Brodu, N.},
  \bibinfo{author}{Leroux, J.}, \bibinfo{year}{2013}.
\newblock \bibinfo{title}{Accurate {3D} comparison of complex topography with
  terrestrial laser scanner: {Application} to the {Rangitikei} {c}anyon
  ({N}-{Z})}.
\newblock \bibinfo{journal}{ISPRS Journal of Photogrammetry and Remote Sensing}
  \bibinfo{volume}{82}, \bibinfo{pages}{10--26}.
\newblock \DOIprefix\doi{10.1016/j.isprsjprs.2013.04.009}.
\bibitem[{Lin et~al.(2021)Lin, Wang and Li}]{lin_2021}
\bibinfo{author}{Lin, D.}, \bibinfo{author}{Wang, J.}, \bibinfo{author}{Li,
  Y.}, \bibinfo{year}{2021}.
\newblock \bibinfo{title}{Unsupervised building damage identification using
  post-event optical imagery and variational autoencoder}.
\newblock \bibinfo{journal}{IEICE Transactions on Information and Systems}
  \bibinfo{volume}{E104.D}, \bibinfo{pages}{1770--1774}.
\newblock \DOIprefix\doi{10.1587/transinf.2021EDL8034}.
\bibitem[{Ma et~al.(2019)Ma, Liu, Zhang, Ye, Yin and Johnson}]{Ma_2019}
\bibinfo{author}{Ma, L.}, \bibinfo{author}{Liu, Y.}, \bibinfo{author}{Zhang,
  X.}, \bibinfo{author}{Ye, Y.}, \bibinfo{author}{Yin, G.},
  \bibinfo{author}{Johnson, B.A.}, \bibinfo{year}{2019}.
\newblock \bibinfo{title}{Deep learning in remote sensing applications: A
  meta-analysis and review}.
\newblock \bibinfo{journal}{ISPRS journal of Photogrammetry and Remote Sensing}
  \bibinfo{volume}{152}, \bibinfo{pages}{166--177}.
\newblock \DOIprefix\doi{10.1016/j.isprsjprs.2019.04.015}.
\bibitem[{Mandlburger et~al.(2017)Mandlburger, Wenzel, Spitzer, Haala, Glira
  and Pfeifer}]{mandlburger_2017}
\bibinfo{author}{Mandlburger, G.}, \bibinfo{author}{Wenzel, K.},
  \bibinfo{author}{Spitzer, A.}, \bibinfo{author}{Haala, N.},
  \bibinfo{author}{Glira, P.}, \bibinfo{author}{Pfeifer, N.},
  \bibinfo{year}{2017}.
\newblock \bibinfo{title}{Improved topographic models via concurrent airborne
  {LiDAR} and dense image matching}.
\newblock \bibinfo{journal}{ISPRS Annals of the Photogrammetry, Remote Sensing
  and Spatial Information Sciences} \bibinfo{volume}{IV-2/W4},
  \bibinfo{pages}{259--266}.
\newblock \DOIprefix\doi{10.5194/isprs-annals-IV-2-W4-259-2017}.
\bibitem[{Meyer and Glock(2022)}]{meyer_2022}
\bibinfo{author}{Meyer, F.}, \bibinfo{author}{Glock, K.}, \bibinfo{year}{2022}.
\newblock \bibinfo{title}{Kinematic orienteering problem with time-optimal
  trajectories for multirotor {UAVs}}.
\newblock \bibinfo{journal}{{IEEE} Robotics and Automation Letters}
  \bibinfo{volume}{7}, \bibinfo{pages}{11402--11409}.
\newblock \DOIprefix\doi{10.1109/lra.2022.3194688}.
\bibitem[{Munawar et~al.(2021)Munawar, Ullah, Qayyum and Heravi}]{munawar_2021}
\bibinfo{author}{Munawar, H.S.}, \bibinfo{author}{Ullah, F.},
  \bibinfo{author}{Qayyum, S.}, \bibinfo{author}{Heravi, A.},
  \bibinfo{year}{2021}.
\newblock \bibinfo{title}{Application of deep learning on uav-based aerial
  images for flood detection}.
\newblock \bibinfo{journal}{Smart Cities} \bibinfo{volume}{4},
  \bibinfo{pages}{1220--1242}.
\newblock \DOIprefix\doi{10.3390/smartcities4030065}.
\bibitem[{Nex et~al.(2019)Nex, Duarte, Tonolo and Kerle}]{nex_2019}
\bibinfo{author}{Nex, F.}, \bibinfo{author}{Duarte, D.},
  \bibinfo{author}{Tonolo, F.G.}, \bibinfo{author}{Kerle, N.},
  \bibinfo{year}{2019}.
\newblock \bibinfo{title}{Structural building damage detection with deep
  learning: Assessment of a state-of-the-art {CNN} in operational conditions}.
\newblock \bibinfo{journal}{Remote Sensing} \bibinfo{volume}{11},
  \bibinfo{pages}{1--17}.
\newblock \DOIprefix\doi{10.3390/rs11232765}.
\bibitem[{North et~al.(2010)North, Rosette, Suárez and Los}]{north_2010}
\bibinfo{author}{North, P.R.J.}, \bibinfo{author}{Rosette, J.A.B.},
  \bibinfo{author}{Suárez, J.C.}, \bibinfo{author}{Los, S.O.},
  \bibinfo{year}{2010}.
\newblock \bibinfo{title}{A monte carlo radiative transfer model of satellite
  waveform {LiDAR}}.
\newblock \bibinfo{journal}{International Journal of Remote Sensing}
  \bibinfo{volume}{31}, \bibinfo{pages}{1343--1358}.
\newblock \DOIprefix\doi{10.1080/01431160903380664}.
\bibitem[{Qin et~al.(2016)Qin, Tian and Reinartz}]{qin_3d_2016}
\bibinfo{author}{Qin, R.}, \bibinfo{author}{Tian, J.},
  \bibinfo{author}{Reinartz, P.}, \bibinfo{year}{2016}.
\newblock \bibinfo{title}{{3D} change detection – {approaches} and
  applications}.
\newblock \bibinfo{journal}{ISPRS journal of Photogrammetry and Remote Sensing}
  \bibinfo{volume}{122}, \bibinfo{pages}{41--56}.
\newblock \DOIprefix\doi{10.1016/j.isprsjprs.2016.09.013}.
\bibitem[{Siddiqui and Awrangjeb(2017)}]{siddiqui_2017}
\bibinfo{author}{Siddiqui, F.U.}, \bibinfo{author}{Awrangjeb, M.},
  \bibinfo{year}{2017}.
\newblock \bibinfo{title}{A novel building change detection method using {3D}
  building models}.
\newblock \bibinfo{journal}{2017 International Conference on Digital Image
  Computing: Techniques and Applications (DICTA)} \bibinfo{volume}{II-3/W5},
  \bibinfo{pages}{1--8}.
\newblock \DOIprefix\doi{10.1109/DICTA.2017.8227394}.
\bibitem[{Stilla and Xu(2023)}]{stilla_2023}
\bibinfo{author}{Stilla, U.}, \bibinfo{author}{Xu, Y.}, \bibinfo{year}{2023}.
\newblock \bibinfo{title}{Change detection of urban objects using 3d point
  clouds: A review}.
\newblock \bibinfo{journal}{ISPRS Journal of Photogrammetry and Remote Sensing}
  \bibinfo{volume}{197}, \bibinfo{pages}{228--255}.
\newblock \DOIprefix\doi{10.1016/j.isprsjprs.2023.01.010}.
\bibitem[{Tran et~al.(2018)Tran, Ressl and Pfeifer}]{tran_2018}
\bibinfo{author}{Tran, T.H.G.}, \bibinfo{author}{Ressl, C.},
  \bibinfo{author}{Pfeifer, N.}, \bibinfo{year}{2018}.
\newblock \bibinfo{title}{Integrated change detection and classification in
  urban areas based on airborne laser scanning point clouds}.
\newblock \bibinfo{journal}{Sensors} \bibinfo{volume}{18},
  \bibinfo{pages}{1--21}.
\newblock \DOIprefix\doi{10.3390/s18020448}.
\bibitem[{{TurboSquid, Inc.}(2023)}]{free3d_2023}
\bibinfo{author}{{TurboSquid, Inc.}}, \bibinfo{year}{2023}.
\newblock \bibinfo{title}{Free3d.} \URLprefix \url{https://free3d.com/}.
\bibitem[{Vetrivel et~al.(2018)Vetrivel, Gerke, Kerle, Nex and
  Vosselman}]{vetrivel_2018}
\bibinfo{author}{Vetrivel, A.}, \bibinfo{author}{Gerke, M.},
  \bibinfo{author}{Kerle, N.}, \bibinfo{author}{Nex, F.},
  \bibinfo{author}{Vosselman, G.}, \bibinfo{year}{2018}.
\newblock \bibinfo{title}{Disaster damage detection through synergistic use of
  deep learning and 3d point cloud features derived from very high resolution
  oblique aerial images, and multiple-kernel-learning}.
\newblock \bibinfo{journal}{ISPRS journal of Photogrammetry and Remote Sensing}
  \bibinfo{volume}{140}, \bibinfo{pages}{45--59}.
\newblock \DOIprefix\doi{10.1016/j.isprsjprs.2017.03.001}.
\bibitem[{Vetrivel et~al.(2015)Vetrivel, Gerke, Kerle and
  Vosselman}]{vetrivel_2015}
\bibinfo{author}{Vetrivel, A.}, \bibinfo{author}{Gerke, M.},
  \bibinfo{author}{Kerle, N.}, \bibinfo{author}{Vosselman, G.},
  \bibinfo{year}{2015}.
\newblock \bibinfo{title}{Identification of damage in buildings based on gaps
  in 3d point clouds from very high resolution oblique airborne images}.
\newblock \bibinfo{journal}{ISPRS journal of Photogrammetry and Remote Sensing}
  \bibinfo{volume}{105}, \bibinfo{pages}{0924--2716}.
\newblock \DOIprefix\doi{10.1016/j.isprsjprs.2015.03.016}.
\bibitem[{Vetrivel et~al.(2016)Vetrivel, Gerke, Kerle and
  Vosselman}]{vetrivel_2016}
\bibinfo{author}{Vetrivel, A.}, \bibinfo{author}{Gerke, M.},
  \bibinfo{author}{Kerle, N.}, \bibinfo{author}{Vosselman, G.},
  \bibinfo{year}{2016}.
\newblock \bibinfo{title}{Potential of multi-temporal oblique airborne imagery
  for structural damage assessment}.
\newblock \bibinfo{journal}{ISPRS Annals of the Photogrammetry, Remote Sensing
  and Spatial Information Sciences} \bibinfo{volume}{III-3},
  \bibinfo{pages}{355--362}.
\newblock \DOIprefix\doi{10.5194/isprs-annals-III-3-355-2016}.
\bibitem[{Winiwarter et~al.(2022)Winiwarter, {Esmorís Pena}, Weiser, Anders,
  {Martínez Sánchez}, Searle and Höfle}]{winiwarter_2022}
\bibinfo{author}{Winiwarter, L.}, \bibinfo{author}{{Esmorís Pena}, A.M.},
  \bibinfo{author}{Weiser, H.}, \bibinfo{author}{Anders, K.},
  \bibinfo{author}{{Martínez Sánchez}, J.}, \bibinfo{author}{Searle, M.},
  \bibinfo{author}{Höfle, B.}, \bibinfo{year}{2022}.
\newblock \bibinfo{title}{Virtual laser scanning with helios++: A novel take on
  ray tracing-based simulation of topographic full-waveform 3d laser scanning}.
\newblock \bibinfo{journal}{Remote Sensing of Environment}
  \bibinfo{volume}{269}, \bibinfo{pages}{1--18}.
\newblock \DOIprefix\doi{10.1016/j.rse.2021.112772}.
\bibitem[{Xu et~al.(2021)Xu, Huang, Song, Ling, Strasbaugh, Yilmaz, Sezen and
  Qin}]{xu_2021}
\bibinfo{author}{Xu, N.}, \bibinfo{author}{Huang, D.}, \bibinfo{author}{Song,
  S.}, \bibinfo{author}{Ling, X.}, \bibinfo{author}{Strasbaugh, C.},
  \bibinfo{author}{Yilmaz, A.}, \bibinfo{author}{Sezen, H.},
  \bibinfo{author}{Qin, R.}, \bibinfo{year}{2021}.
\newblock \bibinfo{title}{A volumetric change detection framework using uav
  oblique photogrammetry – a case study of ultra-high-resolution monitoring
  of progressive building collapse}.
\newblock \bibinfo{journal}{International Journal of Digital Earth}
  \bibinfo{volume}{14}, \bibinfo{pages}{1705--1720}.
\newblock \DOIprefix\doi{10.1080/17538947.2021.1966527}.
\bibitem[{Xu et~al.(2015)Xu, Vosselman and Oude~Elberink}]{xu_2015}
\bibinfo{author}{Xu, S.}, \bibinfo{author}{Vosselman, G.},
  \bibinfo{author}{Oude~Elberink, S.}, \bibinfo{year}{2015}.
\newblock \bibinfo{title}{Detection and classification of changes in buildings
  from airborne laser scanning data}.
\newblock \bibinfo{journal}{Remote Sensing} \bibinfo{volume}{7},
  \bibinfo{pages}{17051--17076}.
\newblock \DOIprefix\doi{10.3390/rs71215867}.
\bibitem[{Zahs et~al.(2022)Zahs, Winiwarter, Anders, Williams, Rutzinger and
  Höfle}]{zahs_2022}
\bibinfo{author}{Zahs, V.}, \bibinfo{author}{Winiwarter, L.},
  \bibinfo{author}{Anders, K.}, \bibinfo{author}{Williams, J.G.},
  \bibinfo{author}{Rutzinger, M.}, \bibinfo{author}{Höfle, B.},
  \bibinfo{year}{2022}.
\newblock \bibinfo{title}{Correspondence-driven plane-based m3c2 for lower
  uncertainty in 3d topographic change quantification}.
\newblock \bibinfo{journal}{ISPRS Journal of Photogrammetry and Remote Sensing}
  \bibinfo{volume}{183}, \bibinfo{pages}{541--559}.
\newblock \DOIprefix\doi{10.1016/j.isprsjprs.2021.11.018}.
\bibitem[{Zhang et~al.(2019)Zhang, Vosselman, Gerke, Persello, Tuia and
  Yang}]{zhang_et_al_2021}
\bibinfo{author}{Zhang, Z.}, \bibinfo{author}{Vosselman, G.},
  \bibinfo{author}{Gerke, M.}, \bibinfo{author}{Persello, C.},
  \bibinfo{author}{Tuia, D.}, \bibinfo{author}{Yang, M.Y.},
  \bibinfo{year}{2019}.
\newblock \bibinfo{title}{Detecting building changes between airborne laser
  scanning and photogrammetric data}.
\newblock \bibinfo{journal}{Remote Sensing} \bibinfo{volume}{11},
  \bibinfo{pages}{1--17}.
\newblock \DOIprefix\doi{10.3390/rs11202417}.
\bibitem[{Zihang et~al.(2015)Zihang, Nagel, Kunde, Hudra, Willkomm, Donabauer,
  Adolphi and Kolbe}]{Zhihang_2018}
\bibinfo{author}{Zihang, Y.}, \bibinfo{author}{Nagel, C.},
  \bibinfo{author}{Kunde, F.}, \bibinfo{author}{Hudra, G.},
  \bibinfo{author}{Willkomm, P.}, \bibinfo{author}{Donabauer, A.},
  \bibinfo{author}{Adolphi, T.}, \bibinfo{author}{Kolbe, T.H.},
  \bibinfo{year}{2015}.
\newblock \bibinfo{title}{3dcitydb - 3d city database version 3.0}.
\newblock \bibinfo{journal}{Open Geospatial Data, Software and Standards}
  \bibinfo{volume}{3}.
\newblock \URLprefix \url{https://www.3dcitydb.org/}.

\end{thebibliography}







\end{document}